\documentclass[lettersize,journal]{IEEEtran}
\usepackage{graphicx}
\usepackage{xcolor}
\usepackage{textcomp}
\usepackage{url}
\usepackage{verbatim}
\usepackage{enumitem}
\usepackage{tikz}

\usepackage{amsmath, amssymb, bm}
\DeclareMathOperator*{\argmin}{arg\,min}
\newcommand{\diag}{\mathrm{diag}}

\usepackage{algorithm}
\usepackage{algorithmic}

\usepackage{acro}
\DeclareAcronym{ICP}{
  short = ICP ,
  long  = iterative closest point,
  tag = acronym
}

\DeclareAcronym{FMCW}{
  short = FMCW ,
  long  = Frequency-Modulated Continuous Wave,
  tag = acronym
}

\DeclareAcronym{lidar}{
  short = lidar,
  long  = Light Detection and Ranging,
  tag = acronym
}

\DeclareAcronym{SOTA}{
  short = SOTA ,
  long  = State-of-the-art,
  list  = State-of-the-Art,
  tag = acronym
}

\DeclareAcronym{TR}{
  short = T\&R ,
  long  = Teach and Repeat,
  tag = acronym
}

\DeclareAcronym{LTR}{
  short = LT\&R ,
  long  = Lidar Teach and Repeat,
  tag = acronym
}

\DeclareAcronym{CDAICP}{
  short = \revision{CDA-ICP},
  long  = \revision{curvature-enhanced degeneracy-aware ICP},
  tag = acronym
}

\DeclareAcronym{GPS}{
  short = GPS ,
  long  = Global Positioning System,
  tag = acronym
}

\DeclareAcronym{RANSAC}{
  short = RANSAC ,
  long  = random sample and consensus,
  list  = Random Sample and Consensus,
  tag = acronym
}

\DeclareAcronym{RMSE}{
  short = RMSE ,
  long  = root mean squared error,
  list  = Root Mean Squared Error,
  tag = acronym
}

\DeclareAcronym{PCA}{
  short = PCA ,
  long  = principal component analysis,
  list  = Principal Component Analysis,
  tag = acronym
}

\DeclareAcronym{WNOA}{
  short = WNOA ,
  long  = white-noise-on-acceleration,
  list  = White-Noise-on-Acceleration,
  tag = acronym
}

\DeclareAcronym{IMU}{
  short = IMU ,
  long  = Inertial Measurement Unit,
  tag = acronym
}

\DeclareAcronym{UGV}{
  short = UGV ,
  long  = unmanned ground vehicle,
  list  = Unmanned Ground Vehicle,
  tag = acronym
}

\DeclareAcronym{GNSS}{
  short = GNSS ,
  long  = Global Navigation Satellite System,
  tag = acronym
}

\DeclareAcronym{OG}{
  short = OG ,
  long  = odometer-gyroscope,
  list  = Odometer-Gyroscope,
  tag = acronym
}

\DeclareAcronym{BCH}{
  short = BCH,
  long = Baker-Campbell-Hausdorff,
  tag = acronym
}

\DeclareAcronym{CSA}{
  short = CSA,
  long = Canadian Space Agency,
  tag = acronym
}

\DeclareAcronym{UTIAS}{
  short = UTIAS,
  long = University of Toronto Institute of Aerospace Studies,
  tag = acronym
}

\DeclareAcronym{DOF}{
  short = DOF,
  long = Degrees of Freedom,
  tag = acronym
}

\DeclareAcronym{SLAM}{
  short = SLAM,
  long = Simultaneous Localization and Mapping,
  tag = acronym
}

\usepackage{array}
\usepackage{booktabs}        
\usepackage{tabularx}        
\usepackage{multirow}
\usepackage{multicol}
\usepackage{makecell}
\usepackage[table]{xcolor}  
\usepackage{pifont}         
\usepackage{stfloats}        

\usepackage[font=small]{caption}
\usepackage[font=small]{subcaption}

\usepackage{cite}

\usepackage{color}

\usepackage{hyperref}

\newcommand{\hide}[1]{}
\begin{document}

\title{\fontsize{22}{28}\selectfont\mdseries 
Degeneracy-Resilient Teach and Repeat for Geometrically \\Challenging Environments Using FMCW Lidar}


\author{Katya M. Papais$^{*}$, Wenda Zhao$^{*}$, Timothy D. Barfoot
\thanks{$^{*}$~Denotes equal contribution.}%
\thanks{The authors are with the Institute for Aerospace Studies, University of Toronto, Canada.}
}

\markboth{IEEE TRANSACTIONS ON ROBOTICS,~Vol.~1,~2026}%
{Shell \MakeLowercase{\textit{et al.}}: A Sample Article Using IEEEtran.cls for IEEE Journals}

\DeclareRobustCommand{\revision}[1]{{\color{black}#1}}
\DeclareRobustCommand{\revisionTodo}[1]{{\color{orange}#1}}
\pdfstringdefDisableCommands{%
  \def\revision#1{#1}%
  \def\revisionTodo#1{#1}%
  \def\ac#1{#1}%
}

\maketitle

\begin{abstract}

Teach and Repeat (T\&R) topometric navigation enables robots to autonomously repeat previously traversed paths without relying on GPS, making it well suited for operations in GPS-denied environments such as underground mines and lunar navigation. State-of-the-art T\&R systems typically rely on iterative closest point (ICP)-based estimation; however, in geometrically degenerate environments with sparsely structured terrain, ICP often becomes ill-conditioned, resulting in degraded localization and unreliable navigation performance. To address this challenge, we present a degeneracy-resilient Frequency-Modulated Continuous-Wave (FMCW) lidar T\&R navigation system consisting of Doppler velocity-based odometry and degeneracy-aware scan-to-map localization. Leveraging FMCW lidar, which provides per-point radial velocity measurements via the Doppler effect, we extend a geometry-independent,  correspondence-free motion estimation to include principled pose uncertainty estimation that remains stable in degenerate environments. We further propose a \revision{curvature-enhanced degeneracy-aware localization method that leverages per-point curvature for improved data association and adaptive registration formulation}, and unifies translational and rotational scales to enable consistent degeneracy detection. \revision{Closed-loop field experiments spanning environments} with varying structural richness demonstrate that the proposed system reliably completes autonomous navigation, including in a challenging flat airport test field where a conventional ICP-based system fails. \revision{We release the implementation of this work at: \href{https://www.ieee-ras.org/about-ras/}{https://opensource\_code/place\_holder}.}



\end{abstract}

\begin{IEEEkeywords}
Field Robots, Localization, State Estimation, Optimization
\end{IEEEkeywords}

\section{Introduction}
\IEEEPARstart{A}{utonomous} navigation in GPS-denied environments, such as underground mines, tunnels, and planetary surfaces, often requires reliable localization without external positioning infrastructure~\cite{marshall2008autonomous, ebadi2023present,daoust2017infrastructure,ingenuity}. In many scenarios, robots are required to perform repeated traversals for cargo transport or inspection between known locations~\cite{dzedzickis2021advanced,krawciw2025sharing,slowik2018automation, LunaCargo}, motivating navigation frameworks that emphasize repeatability, robustness, and efficiency. \ac{TR} topometric navigation is well suited to this setting, as it enables reliable, repeated autonomous traversal of previously driven routes after a single demonstration~\cite{furgale2010visual, krawciw2025local}. By leveraging local submaps constructed during a previously driven path, \ac{TR} avoids the need for global mapping or online terrain assessment and provides consistent path execution, which is especially valuable in GPS-denied and infrastructure-free environments.

\ac{lidar} sensors have proven to be a reliable sensing modality due to their high resolution, accurate range measurements, and invariance to lighting conditions, making \ac{LTR} systems widely adopted for various applications~\cite{baril2022kilometer,krusi2017driving,krawciw2026lunar}. \ac{SOTA} lidar-based estimation is often achieved by geometrically aligning lidar point clouds through an iterative process of nearest-neighbour data association, known as \ac{ICP}-based methods~\cite{besl1992method,pomerleau2015review}. However, many GPS-denied environments, such as flat deserts, open fields, or planetary surfaces, often exhibit large-scale flat and sparsely structured regions, where geometric features are insufficient to fully constrain vehicle motion. In such environments, lidar-based localization methods that rely on ICP registration are prone to geometric degeneracy, leading to unreliable data association and degraded pose estimation. This challenge fundamentally limits the robustness of existing \ac{LTR} systems in geometrically degenerate scenarios.
\begin{figure}[t]
    \centering
    \includegraphics[width=.5\textwidth]{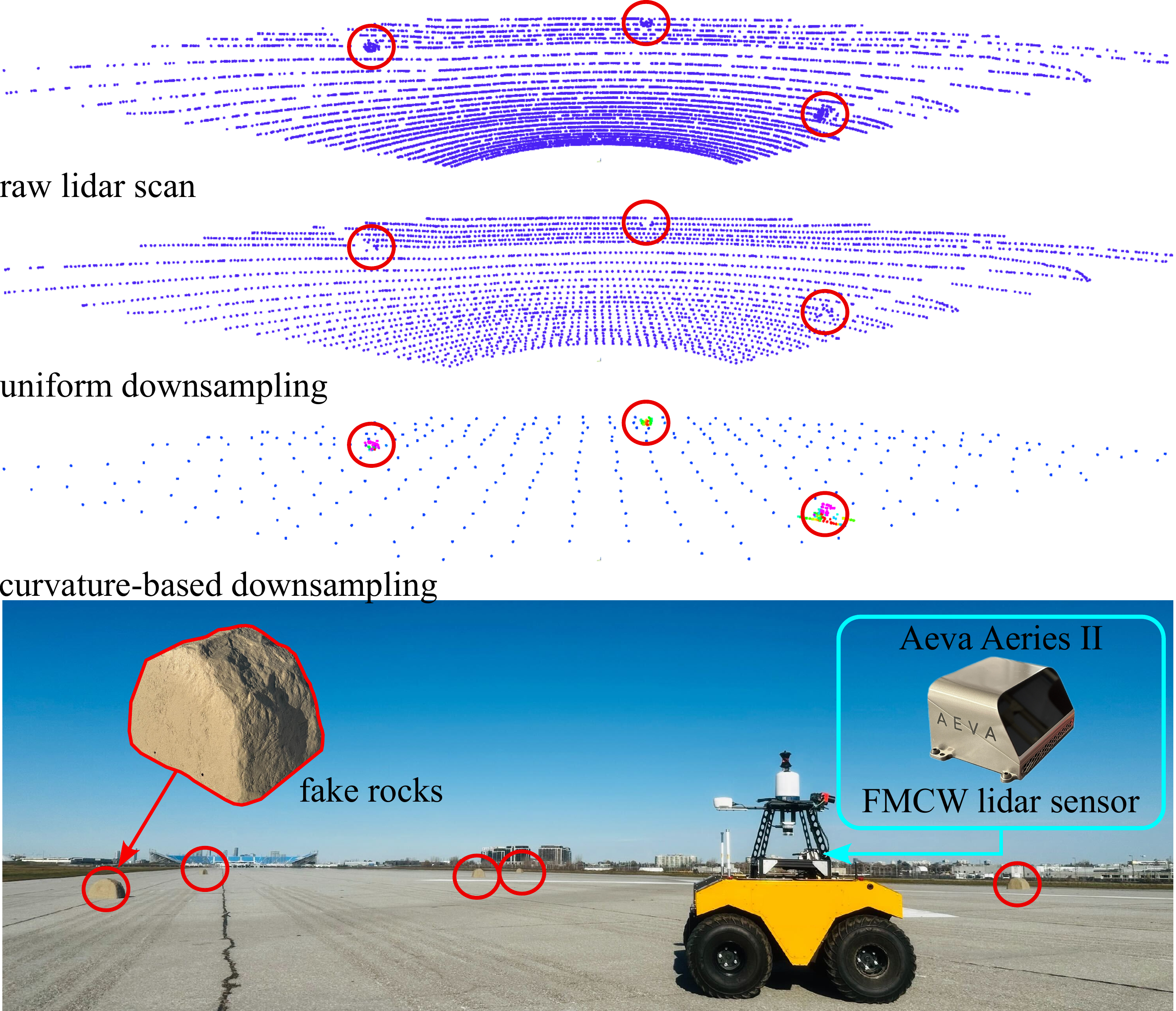}
    \caption{Our proposed degeneracy-resilient FMCW Lidar Teach and Repeat (LT\&R) system operating in a flat airport test field. The top image shows the raw lidar scan, followed by uniformly downsampled and curvature-based downsampled scans (coloured by curvature, where non-planar features are highlighted in different colours). In contrast to uniform downsampling, which treats all points equally and often discards sparse geometric features, curvature-based downsampling selectively reduces flat surfaces and preserves informative geometry, improving computational efficiency and the reliability of scan-to-map data association in geometrically degenerate environments. Fake rocks, circled in red, are added to the scene to introduce minimal geometric structure.}
    \label{fig:airport_image_w_point_cloud}
\end{figure}

Emerging lidar modalities, such as \ac{FMCW} lidar, address this limitation by providing additional motion measurements that are less dependent on geometric structure. This sensing capability has been explored for precise navigation and soft-landing missions on planetary bodies, such as in the Autonomous Landing and Hazard Avoidance (ALHAT) project~\cite{epp2007autonomous, amzajerdian2011lidar}. FMCW lidar employs frequency-domain modulation to estimate range from beat frequency and directly measure per-point radial velocity via the Doppler effect~\cite{pierrottet2008linear,royo2019overview,amzajerdian2012doppler}. These Doppler velocity measurements provide motion information that is particularly valuable in non-distinctive or repetitive environments with insufficient geometric constraints. Consequently, FMCW lidar, also referred to as Doppler lidar, has recently been adopted for Doppler-enhanced lidar odometry in autonomous driving, particularly in scenarios with weak or repetitive geometric structure such as tunnels, highways, and bridges~\cite{Hexsel2022DICPDI,wu2022picking,yoon2023need}.

While Doppler-based odometry provides reliable motion estimates in geometrically degenerate environments, \ac{TR} navigational frameworks require localization relative to the taught path to correct accumulated drift~\cite{krawciw2025local}. In lidar-based systems, this is typically accomplished through scan-to-map registration, which geometrically aligns incoming scans to a locally consistent map. In environments with limited structural variation, however, ICP-based registration becomes ill-conditioned due to insufficient geometric constraints, resulting in unreliable data association and drift along degenerate directions~\cite{gelfand2003geometrically, zhang2016degeneracy, tuna2025informed}. Improving the robustness of scan-to-map localization under such conditions is therefore critical for reliable navigation.

In this paper, we present a degeneracy-resilient \ac{FMCW} \ac{LTR} navigation system. We first extend a Doppler velocity–based correspondence-free lidar–inertial odometry with pose uncertainty estimation, enabling uncertainty propagation from odometry into the localization process. \revision{To improve the robustness of scan-to-map localization, we propose a curvature-enhanced, degeneracy-aware ICP localization algorithm. We leverage per-point curvature to improve data association and adaptively select the appropriate ICP residual based on the local surface geometry. The resulting registration problem is then solved using a degeneracy-aware optimization that unifies translational and rotational scales, enabling more reliable identification of degenerate directions.} \revision{We evaluate the proposed \ac{FMCW} \ac{LTR} system through closed-loop field experiments under diverse structural conditions, ranging from geometrically feature-rich to feature-less scenarios.} In particular, the full system successfully completes autonomous navigation in a flat airport test field (see Figure~\ref{fig:airport_image_w_point_cloud}) where a conventional ICP-based \ac{LTR} system and partially ablated variants fail. Our contributions can be summarized as follows: 
\begin{enumerate}[label=$\bullet$]
  \item We extend a Doppler velocity–based, correspondence-free lidar–inertial odometry framework with pose uncertainty estimation, enabling principled propagation of odometry uncertainty into subsequent localization.
  
  \item We propose a curvature-enhanced, degeneracy-aware ICP algorithm that \revision{leverages} per-point curvature information for improved data association \revision{and adaptive registration formulation,} and unifies translational and rotational scales to enable consistent degeneracy detection.
  
  \item We validate the proposed \ac{FMCW} \ac{LTR} system in closed-loop field experiments across three environments ranging from feature-rich to feature-limited scenarios, and release the implementation to the community.
\end{enumerate}
\section{Related Work}

\subsection{Teach and Repeat Topometric Navigation}
\begin{figure}[t]
    \centering
    \begin{subfigure}[t]{0.48\textwidth}
        \centering
        \includegraphics[width=\textwidth]{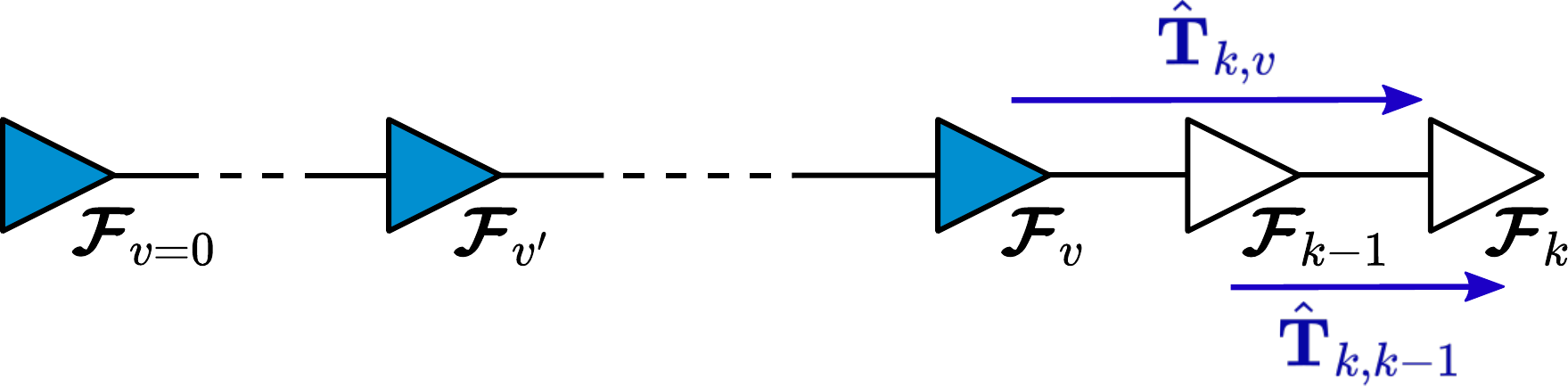}
        \caption{Teach pass}
        \label{fig:teach_pass}
    \end{subfigure}
    \hfill
    \begin{subfigure}[t]{0.45\textwidth}
        \centering
        \includegraphics[width=\textwidth]{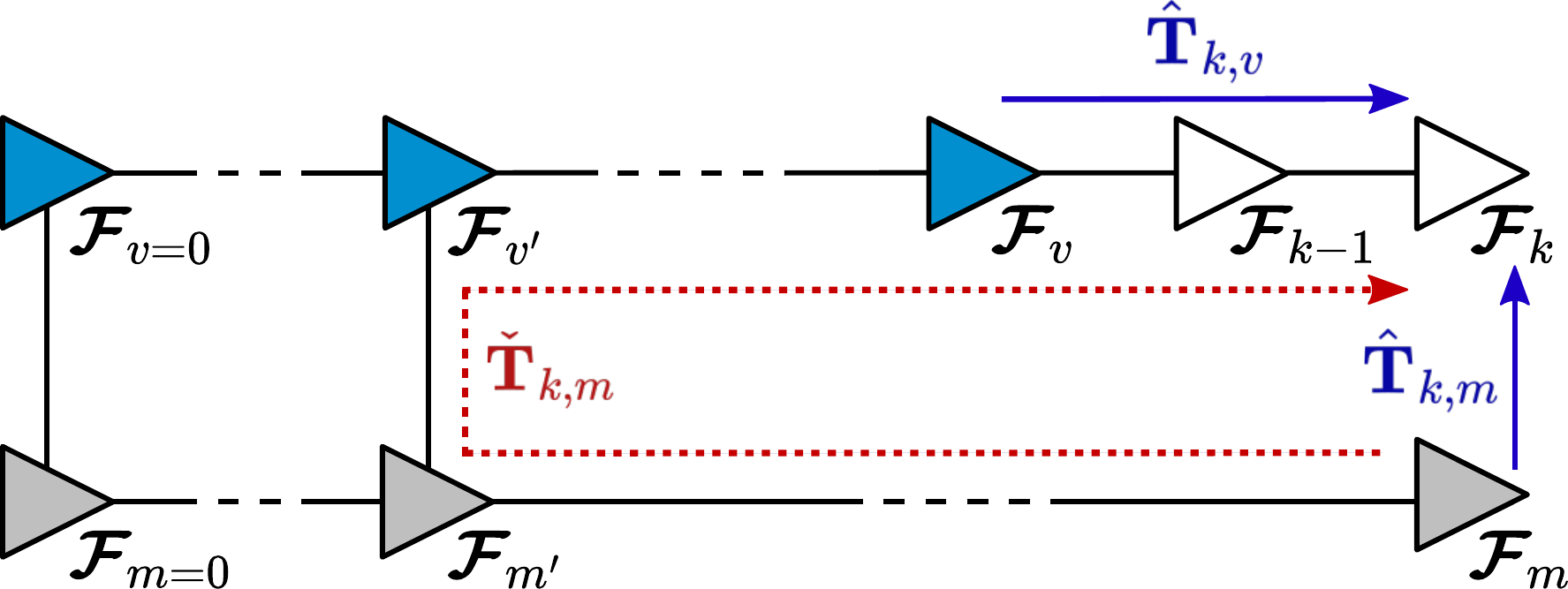}
        \caption{Repeat pass}
        \label{fig:repeat_pass}
    \end{subfigure}
        \caption{Structure of the Teach and Repeat (T\&R) pose graph during (a) the teach pass and (b) the repeat pass. $\boldsymbol{\mathcal{F}}_k$ denotes the current robot frame while $\boldsymbol{\mathcal{F}}_v$ and $\boldsymbol{\mathcal{F}}_m$ indicate the vertex frame from the current pass (teach or repeat), and the local submap frame from the reference (teach) pass, respectively. During both teach and repeat passes, the transformation $\hat{\mathbf{T}}_{k,v}$ from the latest vertex frame $\boldsymbol{\mathcal{F}}_v$ to the robot frame $\boldsymbol{\mathcal{F}}_k$ is estimated using odometry. During the repeat pass, $\boldsymbol{\mathcal{F}}_{v^\prime}$ denotes the most recently localized vertex against the reference vertex $\boldsymbol{\mathcal{F}}_{m^\prime}$. A prior for $\mathbf{T}_{k,m}$ is obtained by compounding transformations through $\boldsymbol{\mathcal{F}}_m$, $\boldsymbol{\mathcal{F}}_{v^\prime}$, $\boldsymbol{\mathcal{F}}_v$, and is refined via scan-to-map localization.  
        }
    \label{fig:t&r_image}
\end{figure}

\ac{TR} navigation is a topometric framework that enables reliable path following by reusing prior driving experience rather than constructing a globally consistent map~\cite{furgale2010visual}. During the teach phase (see Figure~\ref{fig:t&r_image}a), the robot traverses a desired route while building a graph of locally consistent submaps. During the repeat phase, the robot localizes against these submaps to estimate its pose relative to the taught path (as shown in Figure~\ref{fig:t&r_image}b) and track the demonstrated path. By performing localization in local reference frames, \ac{TR} avoids global consistency requirements and remains scalable over long distances~\cite{krawciw2025local}.

A variety of sensing modalities have been explored within the \ac{TR} navigation framework. Early work primarily relied on cameras for metric localization, leading to the widely studied Visual Teach and Repeat (VT\&R) paradigm~\cite{furgale2010visual,churchill2013experience,paton2017can}. However, the \ac{TR} framework generalizes beyond vision, and similar systems have been demonstrated using lidar~\cite{marshall2008autonomous,baril2022kilometer,krusi2017driving}, sonar~\cite{king2018teach}, and radar~\cite{qiao2025radar}. Among these modalities, \ac{LTR} has attracted increasing attention due to its ability to provide accurate geometric measurements that support reliable localization across diverse operational conditions. As a result, \ac{LTR} has been widely adopted in applications such as off-road robotics~\cite{krusi2017driving}, mining~\cite{marshall2008autonomous}, agriculture~\cite{baril2022kilometer}, and planetary rover analogs~\cite{krawciw2026lunar}, where reliable repeated traversal between known locations is required in large-scale and unstructured environments. 

\subsection{Lidar-based Point Cloud Registration}
Lidar-based navigation systems commonly rely on point cloud registration for motion estimation. Registration between consecutive scans, commonly referred to as scan-to-scan registration, primarily provides an odometry estimate. In contrast, scan-to-map registration aligns incoming scans to a locally or globally maintained map and plays a critical role in localization by correcting accumulated odometry drift.

Among the many point-cloud registration techniques proposed in the literature, the \ac{ICP} algorithm remains the most widely adopted approach for lidar-based registration~\cite{pomerleau2015review}. ICP iteratively estimates the rigid-body transformation between a source and a target point cloud by alternating between nearest-neighbour data association and geometric alignment~\cite{besl1992method,dingler2025visual}. A variety of cost functions have been explored, including point-to-point~\cite{besl1992method}, point-to-line~\cite{censi2008icp}, point-to-plane~\cite{chen1992object}, and plane-to-plane~\cite{segal2009generalized}. Other works investigate hybrid, statistically motivated, and certifiable optimization–based formulations~\cite{agamennoni2016point,yokozuka2021litamin2,yang2020teaser}. Despite these alternatives, the point-to-plane cost function~\cite{dannaoui2025and} remains a preferred choice in many \ac{SOTA} lidar-based navigation systems due to its simplicity, robustness, and strong empirical performance in real-world deployments~\cite{carlone2025slam, khattak2020complementary,ebadi2023present}.

The performance of ICP-based registration is strongly influenced by data preprocessing, correspondence quality, and the geometric structure of the point cloud. Standard pipelines typically apply voxel-grid downsampling, normal estimation, and outlier filtering prior to registration~\cite{besl1992method,chen1992object,pomerleau2015review}. While voxel downsampling effectively reduces computational cost~\cite{Burnett2022Ready}, its uniform spatial treatment can disproportionately preserve large planar regions while removing fine geometric details (see Figure~\ref{fig:airport_image_w_point_cloud}, uniform downsampling), which may degrade registration performance in environments with limited structural diversity. 

Several works have emphasized that not all points contribute equally to robust registration. Feature-aware preprocessing and data association strategies have been proposed. Systems such as LOAM~\cite{ZhangLOAM} and its successors, LeGO-LOAM~\cite{LEGOLOAM} and F-LOAM~\cite{wang2021FLOAM}, extract edge and planar points using a local-smoothness score computed along a scan ring, and use different voxel resolutions for the two classes. Surfel-based approaches, such as SuMa~\cite{suma} and SuMa++~\cite{Chen2019SuMaEL}, similarly exploit local normal variation to guide data association. Beyond first-order geometric features, higher-order surface information has also been explored to capture more subtle shape variations. Quadratic surface fitting has been widely used to estimate local curvature, providing a richer description of local geometry~\cite{Khameneifar2018OnTC, Meek2000OnSN, Douros2002TDSCurv}. Gaussian curvature has also been explored within ICP pipelines to assist point selection and correspondence filtering, leveraging its invariance under rigid transformations~\cite{Liang2020CurvReg, Asao2021Curvature}. These studies demonstrate that geometry-aware preprocessing and data association can improve the robustness of lidar-based point cloud registration, particularly in geometrically non-distinctive or repetitive environments.

Our method differs from prior \revision{curvature-enhanced data association} approaches in two main aspects. First, we cluster points into regions of similar curvature and apply separate downsampling to high- and low-curvature clusters, yielding more robust feature preservation in large-scale, geometrically degenerate environments. Second, whereas~\cite{Liang2020CurvReg} relies on RANSAC-based correspondence selection using curvature similarity, we introduce a joint spatial–curvature correspondence metric for ICP registration, which resolves ambiguities in planar or repetitive regions without random sampling (see Figure~\ref{fig:airport_image_w_point_cloud}, curvature-based downsampling).

\subsection{\ac{FMCW} Lidar Odometry}
\begin{figure}[t]
    \centering
    \includegraphics[width=\linewidth]{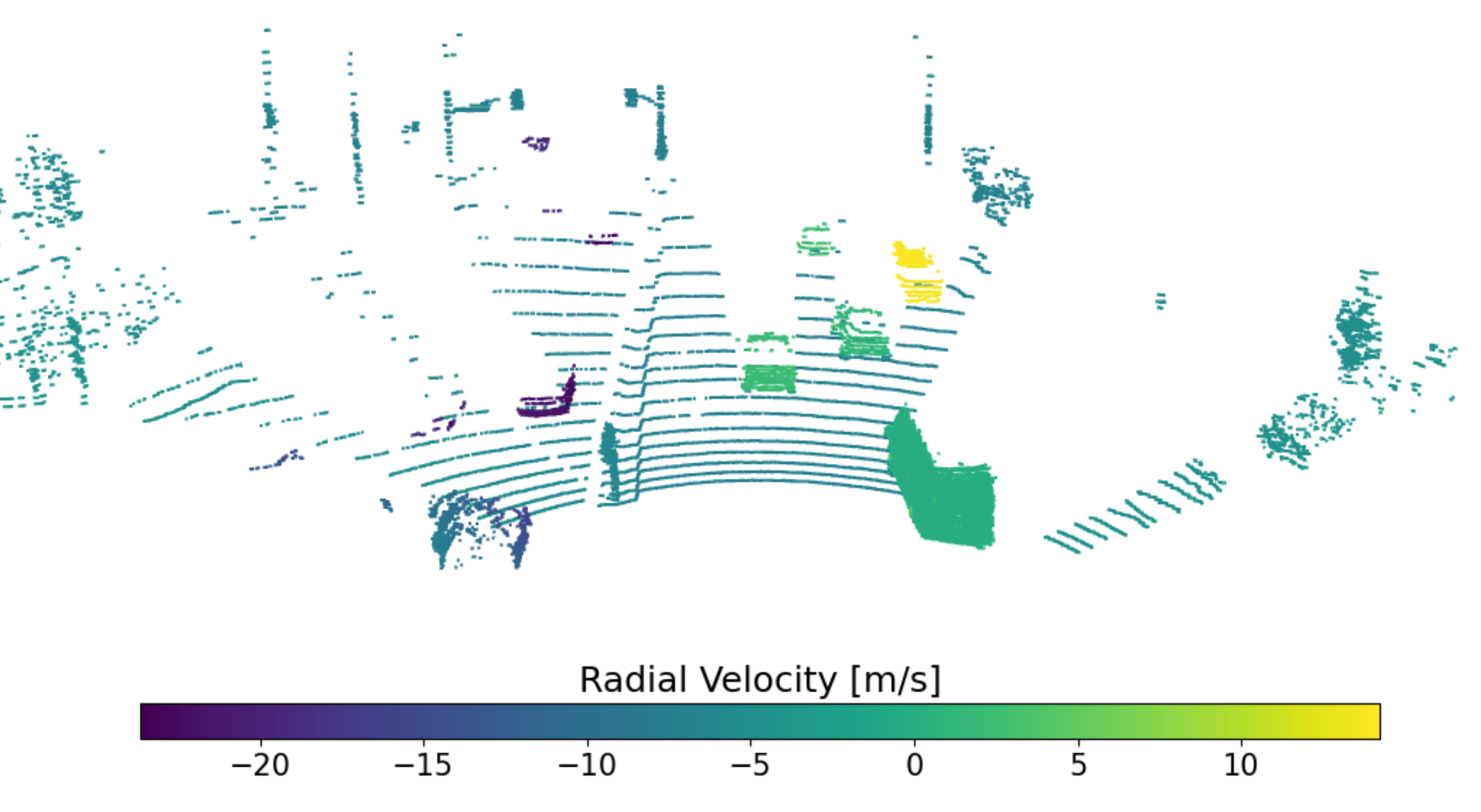}
    \caption{Example \ac{FMCW} lidar scan with points coloured by per-point radial velocity in the lidar frame. Lighter colours indicate positive radial velocity (moving away from the sensor), while darker colours indicate negative radial velocity (moving toward the sensor).}
    \label{fig:FMCW_lidar}
\end{figure}

Recent advances in lidar technology have led to the emergence of \ac{FMCW} lidar sensors, which provide dense 3D point clouds comparable to conventional time-of-flight lidar while additionally measuring per-point relative radial velocity via the Doppler effect~\cite{pierrottet2008linear,royo2019overview,behroozpour2017lidar}. Figure~\ref{fig:FMCW_lidar} illustrates an example \ac{FMCW} lidar point cloud coloured by the Doppler velocity of each point. These properties have motivated growing interest in Doppler velocity–enhanced lidar odometry, particularly in scenarios with weak or repetitive geometric structure, such as tunnels, highways, and bridges.

Hexsel~\textit{et al.}~introduced Doppler-ICP (DICP)~\cite{Hexsel2022DICPDI}, demonstrating that Doppler measurements can be integrated into ICP-based odometry to better constrain motion in environments where geometry alone is insufficient. Zhao~\textit{et al.}~\cite{zhao2024fmcw} integrate Doppler velocity measurements into a lidar–inertial ICP framework to leverage complementary sensing modalities, while Pang~\textit{et al.}~\cite{pang2024efficient} improve computational efficiency through scan-slicing strategies. Wu~\textit{et al.}~\cite{wu2022picking} proposed a continuous-time formulation that directly incorporates Doppler measurements, removing the need for external motion compensation during preprocessing. Building on this idea, Yoon~\textit{et al.}~\cite{yoon2023need} showed that data association between point clouds can be eliminated entirely with a lightweight, correspondence-free Doppler lidar-inertial odometry formulation. By avoiding nearest-neighbour association and explicit geometric matching, this approach is particularly well suited to environments with weak or repetitive geometric structure. Quantitative results show that their method is significantly faster than \ac{SOTA} \ac{ICP}-based odometry while being reasonably accurate. 

Despite these promising results, the correspondence-free Doppler lidar odometry method proposed by Yoon~\textit{et al.}~\cite{yoon2023need} primarily focuses on efficient motion estimation and does not explicitly model or propagate pose uncertainty. Consequently, while the method provides fast and robust odometry estimates in geometrically challenging environments, uncertainty information is not available for subsequent localization or optimization stages, limiting its direct integration into uncertainty-aware navigation frameworks.

\subsection{Localization in Geometrically Degenerate Environments}
There is growing interest in the robotics community in extending robotic systems to operate reliably in challenging, unstructured environments~\cite{ebadi2023present,zhao2025resilient}. \revision{Such environments often contain non-distinctive or repetitive geometric structures that pose significant challenges for \ac{lidar} \ac{ICP}–based navigation systems. Existing solutions commonly} rely on multi-sensor fusion~\cite{zhao2025resilient,zheng2024fast,boche2025okvis2} or the incorporation of complementary information to mitigate lidar degeneracy~\cite{pfreundschuh2024coin, marmaglio2025degeneracy}. \revision{Zhao~\textit{et al.}~\cite{zhao2021super,zhao2025resilient} proposed Super Odometry, which achieves robust odometry in extremely challenging environments through hierarchical adaptation and fusion of complementary sensing modalities. While effective, such approaches rely on additional sensing modalities, motivating continued efforts to improve the robustness of lidar-only pose estimation.} Researchers have also looked into applying adaptive M-estimators~\cite{chebrolu2021adaptive}, better feature sampling~\cite{petracek2024rms}, and adaptive matching techniques~\cite{vizzo2023kiss} to improve robustness of ICP performance. In the autonomous driving literature, dominant ground-plane structures are often explicitly segmented~\cite{liu2019ground,qin2023deep} to improve remote obstacle detection~\cite{zeng2024can} and ICP-based point cloud registration~\cite{wen2022agpc}. However, such strategies typically rely on environment-specific assumptions and are not generally applicable, nor do they fully resolve the underlying optimization degeneracy caused by insufficient geometric constraints. Consequently, mitigating the impact of geometric degeneracy on \ac{ICP} optimization remains fundamentally important for achieving degeneracy-resilient \ac{lidar}-based navigation in the field.


\begin{figure}[t]
    \centering
    \begin{subfigure}[t]{0.48\textwidth}
        \centering
        \includegraphics[width=\textwidth]{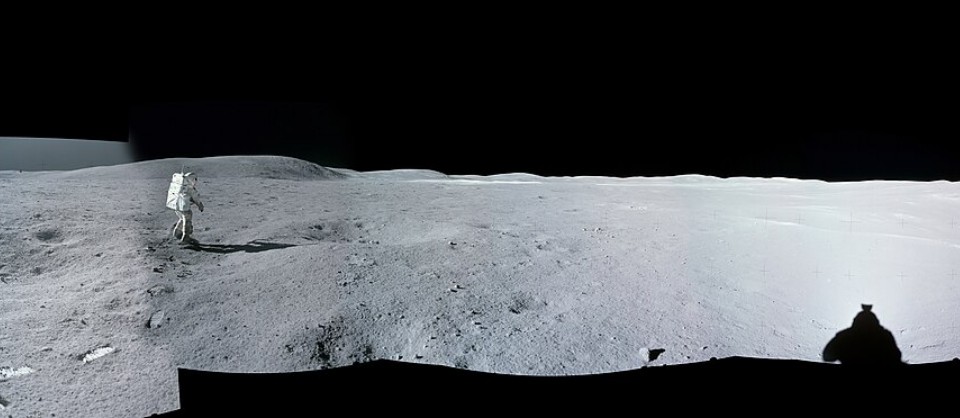}
        \caption{Apollo 16 lunar mission. Courtesy NASA, assembly by Eric Jones~\cite{apollo16}.}
        \label{fig:Apollo_image}
    \end{subfigure}
    \hfill
    \begin{subfigure}[t]{0.48\textwidth}
        \centering
        \includegraphics[width=\textwidth]{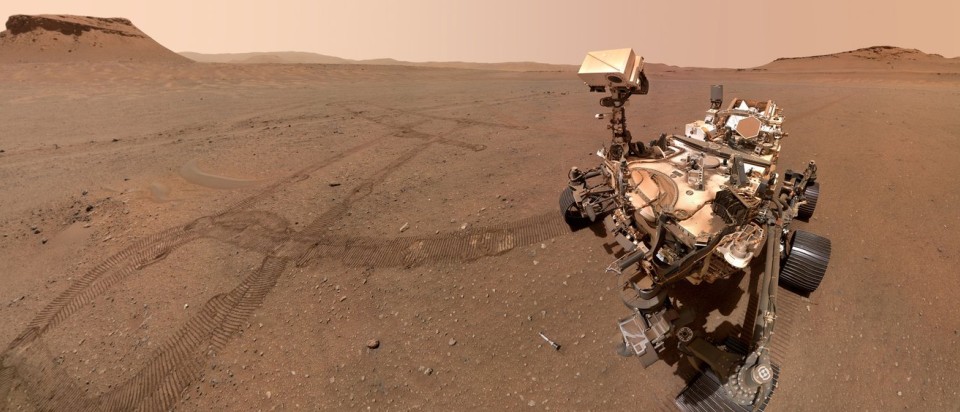}
        \caption{The Perseverance rover on Mars. Courtesy NASA/JPL-Caltech/MSSS~\cite{pereverance}.}
        \label{fig:Mar_image}
    \end{subfigure}
    \caption{(a) Apollo 16 lunar surface imagery (NASA) and (b) Mars 2020 mission imagery (NASA/JPL-Caltech/MSSS) illustrating large-scale, sparsely structured terrain dominated by planar surfaces and gradual elevation changes. In such environments, strong geometric features cannot be assumed, making degeneracy-resilient navigation essential for reliable long-term planetary exploration.}
     \label{fig:aerospace_image}
\end{figure}
The degeneracy of optimization-based \ac{ICP} algorithms has been extensively studied. The pioneering work by Zhang~\textit{et al.}~\cite{zhang2016degeneracy} analytically related the minimal eigenvalue of the Hessian of the optimization to the degeneracy factor and introduced a solution-remapping strategy that updates only the well-conditioned components of the state to handle degenerate cases. Building on this, Hinduja~\textit{et al.}~\cite{hinduja2019degeneracy} employed a Hessian condition number to define an eigenvalue threshold and proposed a degeneracy-aware partial factor for loop-closure in underwater SLAM. These \ac{SOTA} methods have since been widely adopted in robotics, inspiring new extensions~\cite{bai2021degeneration, lee2024switch, yao2025d}. Lee~\textit{et al.}~\cite{lee2024genz} propose to leverage a hybrid point-to-point and point-to-plane error metric to \revision{improve registration robustness in geometrically degenerate environments.} Despite its effectiveness, \revision{scan registration may still suffer from} weakly constrained optimization directions in extremely degenerate environments where planar features dominate (see Figure~\ref{fig:aerospace_image}). 

To explicitly mitigate the lidar ICP degeneracy, Tuna~\textit{et al.}~\cite{tuna2023x} introduce a fine-grained degeneracy detection and impose hard constraints along identified degenerate directions to stabilize the optimization under degeneracy. A follow-up study~\cite{tuna2025informed} provided a comprehensive investigation of degeneracy-resilient techniques for \ac{lidar} localization, and showed that soft-constrained optimization can yield improved performance in complex, poorly conditioned environments when combined with carefully tuned heuristics. To account for the scale difference between translation and rotation elements, the degeneracy detection in~\cite{tuna2023x} and~\cite{tuna2025informed} analyzes translational and rotational Hessian blocks separately and ignores their mutual correlation. However, this separation may lead to inconsistent degeneracy detection in coupled motion scenarios, such as motion relative to a tilted planar surface, where translation and rotation are jointly constrained by the observed geometry. \revision{More recently, Hu~\textit{et al.} proposed DCReg~\cite{hu2025dcreg}, which explicitly accounts for translation-rotation coupling in degeneracy analysis and improves numerical conditioning through structure-aware preconditioning. These recent developments demonstrate the growing importance of robust degeneracy handling in lidar point cloud registration.}

\vspace{6pt}
In this paper, we present a \ac{FMCW} \ac{LTR} system designed to achieve reliable autonomous navigation even in geometrically degraded environments. We first extend a Doppler velocity–based correspondence-free lidar–inertial odometry to include pose uncertainty estimation, enabling uncertainty propagation from odometry into the localization process. To improve scan-to-map localization robustness and efficiency, we leverage Gaussian curvature information extracted from the lidar point cloud. During the teach phase, per-point curvature is computed and stored in local submaps, and subsequently exploited in the repeat phase to enable curvature-aware downsampling and data association. For scan-to-map ICP localization, \revision{we further leverage the Gaussian curvature information to adaptively formulate the point cloud registration objective by selecting the appropriate ICP residual based on the local surface geometry. We then employ} a block-scaling formulation that normalizes rotational and translational components for consistent degeneracy detection, followed by a solution-remapping scheme similar to Hinduja~\textit{et al.}~\cite{hinduja2019degeneracy} to mitigate ill-conditioned updates. To the best of our knowledge, this is the first \ac{FMCW} \ac{LTR} system capable of achieving reliable closed-loop navigation across a wide range of real-world outdoor environments, from highly structured feature-rich areas to severely degenerated feature-less scenarios.

\begin{figure*}[t]
    \centering
    \includegraphics[width=0.95\textwidth]{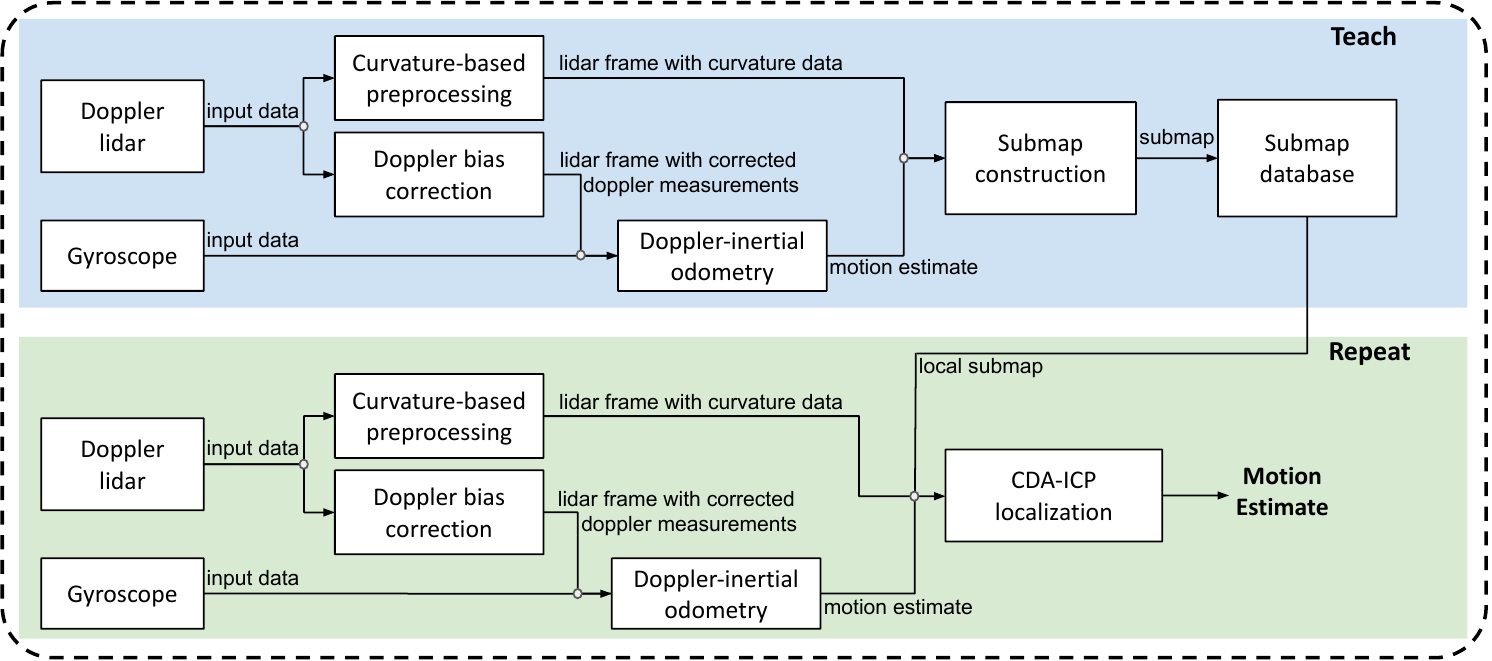}
    \caption{An overview of the proposed Doppler \ac{LTR} system, showing the interaction between modules in the teach and repeat passes. During the teach pass, Doppler velocity and gyroscope measurements are used to estimate correspondence-free Doppler-inertial odometry, referred to throughout as Doppler odometry, while curvature information is computed and stored in local submaps. During the repeat pass, incoming lidar scans are processed using the same curvature-based preprocessing and combined with Doppler odometry to provide a motion prior. Scan-to-map localization is performed using \ac{CDAICP} optimization against the stored submaps to refine the robot pose and enable closed-loop path following in geometrically challenging environments. }
    \label{fig:system-diagram}
\end{figure*}

\section{Methodology}
\subsection{Problem Formulation}
We propose a novel Doppler \ac{LTR} navigation framework that maintains robustness in environments with limited geometric features. The structure of the \ac{TR} pose graph is illustrated in Figure~\ref{fig:t&r_image}, and the overall architecture of our proposed system is illustrated in Figure~\ref{fig:system-diagram}. We use the following notation:
\begin{itemize}
    \item $k$ denotes the current robot frame being estimated,
    \item $v$ denotes the vertex frame from odometry,
    \item $m$ denotes the submap frame associated with vertex $v$ from the teach pass, 
    \item $s$ denotes the lidar sensor frame,
    \item $\mathbf{T}_{m,s}$ denotes the transformation that aligns the live lidar point cloud scan in the sensor frame, $\boldsymbol{\mathcal{F}}_{s}$, to the submap frame, $\boldsymbol{\mathcal{F}}_{m}$,
    \item $\hat{\mathbf{T}}_{k,k-1}$ denotes the latest odometry estimate from the previous frame, $\boldsymbol{\mathcal{F}}_{k-1}$, to the current frame, $\boldsymbol{\mathcal{F}}_{k}$,
    \item $\hat{\mathbf{T}}_{k,v}$ denotes the pose estimate from the vertex frame, $\boldsymbol{\mathcal{F}}_{v}$, to the current robot frame, $\boldsymbol{\mathcal{F}}_{k}$.
\end{itemize}

The goal is to achieve reliable autonomous navigation in geometrically degenerate environments containing sparse features. In both the teach and repeat passes, we use the correspondence-free Doppler–inertial odometry~\cite{yoon2023need}, referred to as Doppler odometry, to estimate the frame-to-frame motion $\hat{\mathbf{T}}_{k,k-1}$ by leveraging the per-point radial velocity and gyroscope measurements from the \ac{FMCW} lidar. To perform localization, we introduce a \revision{curvature-enhanced, degeneracy-aware ICP (\ac{CDAICP})} algorithm that aligns the live lidar scan to the submap and, together with the Doppler odometry prior, provides a robust estimate of $\hat{\mathbf{T}}_{k,v}$. This combination of Doppler odometry and robust \revision{\ac{CDAICP}} localization enables consistent pose estimation and autonomous path following even in geometrically challenging environments where traditional lidar-based methods fail.

\subsection{Curvature-Based Preprocessing} \label{subsec:curv_preprocessing}
A common lidar preprocessing pipeline, such as the method described by Burnett~\textit{et al.}~\cite{Burnett2022Ready}, typically relies on voxel-grid downsampling followed by \ac{PCA}-based plane feature extraction. While effective in structured environments, this approach has limitations in degenerate environments. Uniform voxel downsampling treats all points equally and therefore tends to preserve large planar regions, such as the ground plane, while discarding small, but informative, geometric features. As a result, the downsampled cloud can fail to retain the geometrically distinctive features that support reliable data association, particularly in environments dominated by planar surfaces.

To address this, our preprocessing pipeline adopts a curvature-based downsampling method most similar to the method by~\cite{Liang2020CurvReg}. Rather than uniformly subsampling the cloud, we first compute a per-point curvature estimate that captures local surface variation. For each point, a $k$-nearest neighbour search is performed using a KD-tree, and \ac{PCA} is applied to obtain an initial estimate of the surface normal and principal directions. Since \ac{PCA} yields surface normals up to a sign ambiguity, the normal orientation is disambiguated by enforcing consistency with the lidar viewpoint, ensuring that normals face the sensor origin. The point neighbourhood is then expressed in a local tangent frame and fitted with a quadratic surface of the form

\begin{equation}
    z(x,y) = ax^2 + bxy + cy^2 + dx + ey + f.
\end{equation}
The coefficients of this fit yield the first and second fundamental forms, from which curvature, $\kappa$, is computed as
\begin{equation}
    \kappa = \frac{ln - m^2}{eg - f^2},
\end{equation}
where $(l, m, n)$ and $(e, f, g)$ denote the components of the second and first fundamental forms, respectively. This quadratic fit refines the curvature estimate beyond what \ac{PCA} alone can compute, enabling the detection of smaller geometric differences such as ridges or transitions between different surface patches (e.g., asphalt versus grass).

The point cloud is then clustered into regions that share a similar curvature value. First, points with curvature below a threshold are classified as approximately planar and are grouped into a single cluster. Non-planar points are then clustered via a union--find region-growing procedure: each point searches for nearby neighbours within a fixed radius, and clusters are formed by merging connected sets of non-flat points. Small clusters, typically from noise or outliers, are removed based on a minimum cluster size criterion.

\begin{figure*}[t!]
    \centering
    \includegraphics[width=\textwidth]{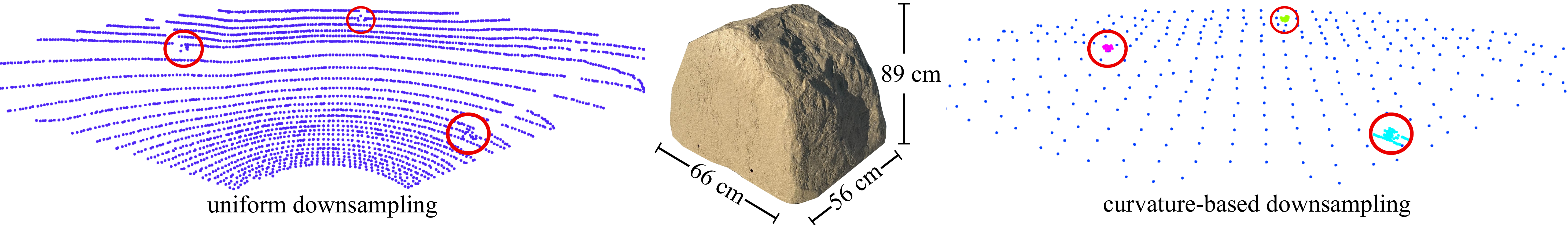}
    \caption{Uniform downsampling (left) removes points indiscriminately across the scene, while curvature-based downsampling (right) preserves geometric features with high curvature, such as the rocks, and reduces redundant planar points. The curvature-based downsampled frame is coloured by curvature cluster.}
    \label{fig:downsample_comparison}
\end{figure*}

This clustering enables selective, geometry-aware downsampling. Each cluster is processed independently, and its mean curvature determines the voxel size used. Low-curvature clusters (e.g., ground plane, building facades) are aggressively downsampled using a larger voxel size, reducing redundant data in areas with low information. High-curvature clusters, which contain the most informative points for scan registration, are downsampled at a smaller resolution to preserve detail even when these features are sparse. An example of curvature-based point cloud downsampling, using a scene containing three artificial rocks placed on a flat airport runway, is illustrated in Figure~\ref{fig:downsample_comparison}. The resulting point cloud retains a dense sampling of the geometrically rich regions, while significantly reducing redundant planar points.

\revision{
\begin{algorithm}[t]
    \caption{Curvature-Based Adaptive Point Cloud Downsampling}
    \label{alg:curvature}
    \begin{algorithmic}[1]
    \REQUIRE Point cloud $P$; max points $N_{\max}$; planarity ratio $\rho$; curvature threshold $\kappa_{\mathrm{th}}$; cluster radius $r_c$; voxel sizes $v_{p}$ (planar), $v_{f}$ (feature)
    \ENSURE Downsampled point cloud $P_{\mathrm{out}}$
    
    \STATE Crop $P$ to sensing range; if $\text{card}(P) > N_{\max}$, randomly subsample to $N_{\max}$
    
    \FOR{each point $p_i \in P$}
        \STATE Compute eigenvalues $\lambda_0 \le \lambda_1 \le \lambda_2$ from the local covariance matrix
        \IF{$\lambda_0 / \lambda_1 < \rho$}
            \STATE $\kappa(p_i) \leftarrow 0$ \COMMENT{approximately planar}
        \ELSE
            \STATE Fit a local quadratic surface; compute $\kappa(p_i)$
        \ENDIF
    \ENDFOR
    
    \STATE Partition $P$ into planar set $P_{\mathrm{flat}}$ ($|\kappa| < \kappa_{\mathrm{th}}$) and non-planar set $P_{\mathrm{feat}}$
    \STATE Cluster $P_{\mathrm{feat}}$ via radius-$r_c$ union--find
    
    \FOR{each cluster $C$}
        \STATE $\bar{\kappa}_C \leftarrow \frac{1}{|C|}\sum_{p_i \in C} \kappa(p_i)$ \COMMENT{mean curvature of $C$}
        \IF{$|\bar{\kappa}_C| < \kappa_{\mathrm{th}}$}
            \STATE Voxel-downsample $C$ with $v_{p}$
        \ELSE
            \STATE Voxel-downsample $C$ with $v_{f}$
        \ENDIF
    \ENDFOR
    
    \RETURN $P_{\mathrm{out}} \leftarrow$ union of all downsampled clusters
\end{algorithmic}
\end{algorithm}

%
    
    
    
    
    

Algorithm~\ref{alg:curvature} summarizes the procedure described above. The overall complexity is $\mathcal{O}(N\log N)$, dominated by the KD-tree construction and $k$-nearest-neighbour searches used for curvature estimation and clustering, while the quadratic surface fit and union--find clustering each contribute at most $\mathcal{O}(N)$. The curvature estimation and clustering stages are parallelized across CPU threads using OpenMP, with resulting per-scan runtimes reported in Section~\ref{subsec:runtime}. 
} 

\subsection{Doppler Odometry}
As we target navigating in environments with non-distinctive or repetitive geometric environments, we adopt the Doppler odometry method presented by Yoon~\textit{et al.}~\cite{yoon2023need}, a geometry-independent and correspondence-free odometry estimator. We provide a brief summary and refer the reader to~\cite{yoon2023need} for more detail. 

The state of the Doppler odometry only contains the 6-DOF vehicle body velocities $\bm{\varpi}(t)$, where $\bm{\varpi}$ is a $6\times 1$ vector containing the body-centric linear and angular velocities. The continuous-time estimation framework of Barfoot~\textit{et al.}~\cite{barfoot2014batch} is applied to estimate the trajectory as a Gaussian process (GP), which allows each measurement to be applied at \revision{its} exact measurement time. The relative pose estimate is computed via numerical integration. 

The vehicle velocity prior is modelled as White-Noise-on-Acceleration (WNOA)
\begin{equation}
    \dot{\bm{\varpi}}(t) = \mathbf{w}(t), ~~ \mathbf{w}(t)\sim \mathcal{GP}(\mathbf{0}, \mathbf{Q}_c \delta(t-t')),
\end{equation}
where $\mathbf{Q}_c$ is the power spectral density matrix. Incorporating the vehicle kinematics by penalizing velocities in the lateral, vertical, roll, and pitch dimensions, the kinematic residual term is
\begin{equation}
    \mathbf{e}_{\text{kin},k} = \mathbf{H}_\revision{\textrm{kin}} \boldsymbol{\varpi}_{k},
\end{equation}
where a constant $\mathbf{H}_{\textrm{kin}}$ extracts the dimensions of interest.

The residual model for Doppler measurements is 
\begin{equation}
    e_{\text{dop},i} = y_{\text{dop},i} - \frac{1}{||\mathbf{q}_{i}||} \left[ \mathbf{q}^{\top}_{i} \hspace{0.2cm} \mathbf{0}\revision{_{1\times 3}} \right] \boldsymbol{\mathcal{T}}_{s,r} \boldsymbol{\varpi}(t_{i}) - h(\boldsymbol{\psi}_{i}),
\end{equation}
where $y_{\text{dop},i}$ is the $i\text{th}$ Doppler measurement, $ \mathbf{q}_{i} \in \mathbb{R}^{3}$ are the corresponding point coordinates in the sensor frame, $\boldsymbol{\mathcal{T}}_{s,r} = \textrm{Ad}(\mathbf{T}_{s,r})$ is the known extrinsic adjoint transformation between the sensor and robot, \revision{and $h(\bm{\psi}_i)$ is the Doppler velocity bias calibrated offline using the linear regression model from~\cite{yoon2023need}.} This error measures the discrepancy observed between the measured radial velocity and the predicted radial velocity induced by the vehicle's motion. Due to the continuous-time formulation, the body-centric velocity can be conveniently queried using linear interpolation~\cite{barfoot2014batch} as

\begin{equation} \label{eqn:interpolation}
    \bm{\varpi}(\tau) = (1-\alpha)\bm{\varpi}_{k-1} + \alpha\bm{\varpi}_{k}, ~~~ \alpha = \frac{\tau-t_{k-1}}{t_{k}-t_{k-1}},
\end{equation}
with $\tau \in [t_{k-1},t_{k}]$. Therefore, $\bm{\varpi}(t_i)\in\mathbb{R}^6$ is the vehicle velocity queried at the lidar sensor measurement time $t_i$.

As the angular velocity is unobservable with Doppler measurements from one \ac{FMCW} sensor~\cite{yoon2023need}, gyroscope data is used to estimate the angular velocity. The residual model is 
\begin{equation} \label{eqn:gyro_error}
    \mathbf{e}_{\text{gyro},j} = \mathbf{y}_{\text{gyro},j} - \mathbf{R}_{s,r}^{} \mathbf{D} \boldsymbol{\varpi}(t_{j}) - \mathbf{b}_{\textrm{gyro}},
\end{equation}
where $\mathbf{y}_{\text{gyro},j} \in \mathbb{R}^3$ is the $j\textrm{th}$ angular velocity measurement in the sensor frame, and $\mathbf{R}_{s,r}\in$ $SO(3)$ is the known extrinsic rotation between the sensor and robot. The constant $3\times 6$ projection matrix $\mathbf{D}$ removes the translational elements of body-centric velocity $\bm{\varpi}(t_j)$ queried at timestamp $t_j$. Since the gyroscope bias $\mathbf{b}_{\textrm{gyro}}$ is reasonably constant in practice, we compute the value of $\mathbf{b}_{\textrm{gyro}}$ by averaging the gyroscope's angular velocity measurements over 30 seconds while the robot is stationary. During the repeat pass, the gyroscope bias is updated using the method described in the \revision{supplementary material}\footnote{\href{https://www.ieee-ras.org/about-ras/}{https://supplementary\_material/place\_holder}\label{fn:supplementary}}.

With the motion prior and the Doppler and gyroscope measurements, the odometry is formulated as linear continuous-time batch estimation using a Maximum A Posteriori (MAP) problem
\begin{equation}
    J_{\text{odom}} = \sum_k ( \phi_{\text{wnoa},k} + \phi_{\text{kin},k}) + \sum_i \phi_{\text{dop},i} + \sum_j \phi_{\text{gyro},j},
\end{equation}
where the WNOA motion prior factor $\phi_{\textrm{wnoa}, k}$, vehicle kinematic factor $\phi_{\textrm{kin}, k}$, Doppler measurement factor $\phi_{\textrm{dop}, i}$, and gyroscope measurement factor $\phi_{\textrm{gyro}, j}$ are 
\begin{equation}
    \begin{split}
        \phi_{\textrm{wnoa}, k} &= \frac{1}{2}(\bm{\varpi}_k - \bm{\varpi}_{k-1})^\top \mathbf{Q}_k^{-1} (\bm{\varpi}_k - \bm{\varpi}_{k-1}), \\
        \phi_{\textrm{kin}, k} &= \frac{1}{2}{\mathbf{e}_{\textrm{kin},k}}^\top \mathbf{Q}_z {\mathbf{e}_{\textrm{kin},k}}, \\
        \phi_{\textrm{dop}, i} &= \frac{1}{2}(e_{\textrm{dop}, i})^2 R^{-1}_{\textrm{dop}}, \\
        \phi_{\textrm{gyro}, j} &= \frac{1}{2}{\mathbf{e}_{\textrm{gyro}, j}}^\top \mathbf{R}_{\textrm{gyro}}^{-1} {\mathbf{e}_{\textrm{gyro}, j}}.
    \end{split}
\end{equation}
We solve the joint optimization problem and incrementally marginalize out the past velocity states to compute the latest velocity~$\bm{\varpi}_{k}$ (e.g., a filter implementation that handles measurements asynchronously). The relative pose estimate is approximated by numerically integrating: 
\begin{equation} \label{eqn:pose_num_int}
    \begin{split}
        \hat{\mathbf{T}}_{k, k-1} \approx \prod^{S}_{i=1} \exp(\Delta t \boldsymbol{\varpi}(t_{k-1} + i \Delta t)^{\wedge}),
    \end{split}
\end{equation}
where $S$ is the number of interpolation steps, $\text{exp}(\cdot)$ is the exponential map, and $(\cdot)^{\wedge}$ transforms an element of $\mathbb{R}^6$ into a member of Lie algebra $\mathfrak{se}(3)$~\cite{barfoot2024state}.

In addition to estimating the vehicle velocity and relative poses, we derived the associated pose uncertainty not presented in~\cite{yoon2023need}, which allows us to propagate the uncertainties from the Doppler odometry estimates into the localization step.

\paragraph{Covariance of Interpolated Velocities}
The velocity covariance at the query time $\tau$, denoted $\mathbf{P}(\bm{\varpi}_\tau, \bm{\varpi}_\tau)$, is obtained by combining the batch-estimated covariance between consecutive frame velocities with the integrated process noise. Specifically, from the batch estimation we have the joint covariance
\begin{equation}
    \mathbf{P}(\bm{\varpi}_{t_{k-1}}, \bm{\varpi}_{t_{k}}) = \mathbf{A}_{\textrm{odom}}^{-1},
\end{equation}
where $\mathbf{A}_{\textrm{odom}}$ is the Hessian matrix of the Doppler odometry least-squares problem, and the inverse arises naturally from solving the linear system $\mathbf{A}_{\textrm{odom}}\mathbf{x} = \mathbf{b}_{\textrm{odom}}$ with $\mathbf{x} = [\bm{\varpi}_{t_{k-1}}^\top \bm{\varpi}_{t_{k}}^\top]^\top $.  

To query the covariance at an intermediate time $\tau \in [t_{k-1}, t_{k}]$, we interpolate using the Gaussian process framework. Following the continuous-time WNOA model, the mean velocity is linearly interpolated with the interpolation factor from \eqref{eqn:interpolation}.
The corresponding interpolated covariance is
\begin{equation}
\begin{aligned}
\mathbf{P}(\bm{\varpi}_\tau, \bm{\varpi}_\tau)
    &= 
    \begin{bmatrix} (1-\alpha)\mathbf{1}_6 & \alpha\mathbf{1}_6 \end{bmatrix}
    \mathbf{P}(\bm{\varpi}_{t_{k-1}}, \bm{\varpi}_{t_k})
    \begin{bmatrix} (1-\alpha)\mathbf{1}_6 \\ \alpha\mathbf{1}_6 \end{bmatrix}
    \\
    &\quad + (1-\alpha)(\tau - t_{k-1})\mathbf{Q}_c .
\end{aligned}
\end{equation}
following~\cite{anderson2017batch}.

The first term propagates the uncertainty from the batch-estimated velocities, while the second term accounts for the accumulated process noise between $t_{k-1}$ and $\tau$ under the WNOA prior. This formulation ensures that both correlations between frame velocities and the continuous-time process noise are captured in the interpolated covariance.

\paragraph{Numerical Pose Integration with Uncertainty}

To compute the relative pose between frames, we numerically integrate the interpolated body-centric velocities over small timesteps $\Delta t$ using the incremental update, described in \eqref{eqn:pose_num_int}. We now derive how uncertainty in these queried velocities propagates through each integration step.

We represent the pose at the previous timestep using the standard perturbation model:
\begin{equation}
    \hat{\mathbf{T}}_{k-1} = \exp(\boldsymbol{\epsilon}_{k-1}^{\wedge})\,\mathbf{T}_{k-1}, 
    \qquad 
    \boldsymbol{\epsilon}_{k-1} \sim \mathcal{N}(\mathbf{0}, \mathbf{\Sigma}_{k-1}).
\end{equation}
To understand how these uncertainties combine, we consider one integration step to the pose estimation at time $\tau$, where we propagate the uncertainty
\begin{equation}
\begin{aligned}
    \hat{\mathbf{T}}_{\tau}
    &= \exp\!\left( (\bm{\phi}+\delta\bm{\phi})^{\wedge} \right)\,
       \mathbf{T}_{k-1}.
  \end{aligned}
\end{equation}
We first isolate the effect of the velocity noise. Using the first-order left Jacobian perturbation identity
\begin{equation}
    \exp\!\left( (\bm{\phi}+\delta\bm{\phi})^{\wedge} \right)
    \approx
    \exp\!\left( (\boldsymbol{\mathcal{J}}_l(\bm{\phi})\,\delta\bm{\phi})^{\wedge} \right)
    \exp(\bm{\phi}^{\wedge}),
\end{equation}
we separate the nominal increment from its perturbation \cite{barfoot2024state} by using the $SE(3)$ left Jacobian, $\boldsymbol{\mathcal{J}}_l(\bm{\phi})$.

Next, we incorporate the uncertainty in the previous pose.  
Substituting $\hat{\mathbf{T}}_{k-1}=\exp(\boldsymbol{\epsilon}_{k-1}^{\wedge})\mathbf{T}_{k-1}$
and we can transport the pose perturbation \cite{barfoot2024state} by
\begin{equation}
\begin{aligned}
    \hat{\mathbf{T}}_{\tau}
    \approx ~
    &\exp\!\left( (\boldsymbol{\mathcal{J}}_l(\bm{\phi})\delta\bm{\phi})^{\wedge} \right) \\
    & \exp\!\left( (\text{Ad}(\exp(\bm{\phi}^{\wedge}))\, \boldsymbol{\epsilon}_{k-1})^{\wedge} \right)
    \exp(\bm{\phi}^{\wedge})\mathbf{T}_{k-1},
\end{aligned}
\end{equation}
where $\text{Ad}(\cdot)$ is the adjoint operator of a Lie group. 

The product of the two perturbations is now expressed in the Lie algebra. Applying the first-order \ac{BCH} expansion, we have
\begin{equation}
\begin{aligned}
    \bm{\xi}_{\tau}
    &= 
    \ln\!\left(
      \exp\!\left((\boldsymbol{\mathcal{J}}_l(\bm{\phi})\delta\bm{\phi})^{\wedge}\right)
      \exp\!\left((\text{Ad}(\exp(\bm{\phi}^{\wedge}))\,\boldsymbol{\epsilon}_{k-1})^{\wedge}\right)
    \right)^{\vee} \\
    &\approx 
    \boldsymbol{\mathcal{J}}_l(\bm{\phi})\,\delta\bm{\phi}
    +
    \text{Ad}(\exp(\bm{\phi}^{\wedge}))\,\boldsymbol{\epsilon}_{k-1},
\end{aligned}
\end{equation}
which expresses the stochastic increment, $\bm{\xi}_\tau$, as the sum of the propagated velocity uncertainty and the transported prior pose uncertainty.

The covariance of this incremental update is therefore
\begin{equation}
\begin{split}
    \mathbf{P}(\hat{\mathbf{T}}_{\tau})
    &=\text{Cov}\left(\bm{\xi}_{\tau}\right) \\
    &= \boldsymbol{\mathcal{J}}_l(\bm{\phi})\textrm{Cov}(\bm{\phi})\boldsymbol{\mathcal{J}}_l(\bm{\phi})^\top + \\
    &~~~~~~~~~~~~~~ \text{Ad}(\exp(\bm{\phi}^{\wedge})) \textrm{Cov}(\boldsymbol{\epsilon}_{k-1})\text{Ad}(\exp(\bm{\phi}^{\wedge}))^\top \\
    &= \Delta t^{2}\,
    \boldsymbol{\mathcal{J}}_l(\bm{\phi})\,
    \mathbf{P}(\bm{\varpi}_\tau,\bm{\varpi}_\tau)\,
    \boldsymbol{\mathcal{J}}_l(\bm{\phi})^{\top}
    + \\
    &~~~~~~~~~~~~~~ \text{Ad}(\exp(\bm{\phi}^{\wedge}))\,
    \mathbf{\Sigma}_{k-1}\,
    \text{Ad}(\exp(\bm{\phi}^{\wedge}))^{\top}.
\end{split}
\end{equation}
This expression is applied recursively across all integration steps to yield the full pose uncertainty at time $k$.

\paragraph{Recursive Pose Covariance Update}
By applying this update recursively for all integration steps within a frame, we compute the mean pose and associated covariance at the end of the lidar frame. This allows the Doppler odometry estimator to maintain a consistent estimate of both pose and uncertainty:
\begin{equation}
    \hat{\mathbf{T}}_{k} = \hat{\mathbf{T}}_{k,k-1} \hat{\mathbf{T}}_{k-1}, \quad
    \mathbf{\Sigma}_{k} = \mathbf{P}(\hat{\mathbf{T}}_{k}).
\end{equation}

\subsection{Mapping} \label{subsec:wart_mapping}
In the \ac{TR} framework, mapping is achieved by incrementally building a series of local submaps that represent the vehicle's environment. Using the odometry estimates, the system continuously monitors the vehicle's motion, estimating the transformation between the latest submap, $\boldsymbol{\mathcal{F}}_{m}$, and the current live lidar scan, $\boldsymbol{\mathcal{F}}_{k}$. When the relative motion exceeds a predefined translation or rotation threshold, a new submap vertex $\boldsymbol{\mathcal{F}}_{m+1}$ is added.

The reference maps are constructed during the teach pass using the Doppler odometry described above. We compute and store the curvature information of each point in a lidar frame using the curvature preprocessing method presented in Section~\ref{subsec:curv_preprocessing}; however, during the teach pass, we do not apply the feature-specific downsampling. Instead of separately downsampling low- and high-curvature features, we apply the small feature voxel size to the full point cloud (equivalent to the preprocessing method presented in~\cite{Burnett2022Ready}). This retains a high point density in the stored submaps while preserving the curvature attributes required for feature matching during the repeat pass.

\subsection{\revision{Curvature-Enhanced,} Degeneracy-Aware ICP Localization} \label{subsec:wart_daicp_loc}
During the repeat pass, localization is performed by registering incoming scans to the previously constructed reference submaps using a \revision{curvature-enhanced, degeneracy-aware ICP~(\ac{CDAICP})} algorithm. \revision{We leverage the estimated Gaussian curvature to improve correspondence search and to adaptively determine the geometric registration constraint for each matched correspondence.}

\subsubsection{Data Association} \label{sss:wart_da}
To improve robustness in geometrically degenerate environments, we augment the standard spatial nearest-neighbour \revision{search} to include curvature information in the point correspondence search. This ensures that points are paired not only based on Euclidean proximity but also on local surface structure, reducing incorrect associations on planar regions that otherwise have ambiguous nearest neighbours.

Let $\mathcal{P}^{\textrm{raw}}=\{\mathbf{p}_i^{\textrm{raw}}\}_{i=1}^{N_p}$ and $\mathcal{Q}^{\textrm{raw}}=\{\mathbf{q}_j^{\textrm{raw}}\}_{j=1}^{N_q}$ denote the raw lidar scan and submap point clouds. For each scan point $\mathbf{p}^{\textrm{raw}}\in\mathcal{P}^{\textrm{raw}}$, we denote $\mathcal{K}\left(\mathbf{p}^{\textrm{raw}}\right) \subset \mathcal{Q}^{\textrm{raw}}$ as the set of its $k$-nearest map neighbours, with $k=8$. In this work, we propose using a joint spatial-curvature score that combines Euclidean proximity and curvature similarity to evaluate candidate map points $\mathbf{q}^{\textrm{raw}}\in \mathcal{K}\left(\mathbf{p}^{\textrm{raw}}\right)$ for point correspondence. The correspondence score is defined as
\begin{equation}
    c(\mathbf{p}^{\textrm{raw}}, \mathbf{q}^{\textrm{raw}}) = 
    \frac{|\kappa_p - \kappa_q|}{\eta_\kappa} 
    + \beta \frac{\|\mathbf{p}^{\textrm{raw}} - \mathbf{q}^{\textrm{raw}}\|}{\eta_d},
\end{equation}
where $\kappa_p$ and $\kappa_q$ denote point curvature corresponding to points $\mathbf{p}^{\textrm{raw}}$ and $\mathbf{q}^{\textrm{raw}}$, $\eta_\kappa$ and $\eta_d$ are normalization scales that convert curvature (units $\mathrm{m}^{-1}$) and Euclidean distance (units $\mathrm{m}$) into comparable dimensionless quantities, and a weighting parameter $\beta$ balances the influence of geometric similarity and spatial proximity. The optimal correspondence is selected as the candidate that minimizes this score over the neighbour set.

Rather than assuming fixed normalization constants, the scaling terms are estimated online using pseudo-random sampling of up to $1000$ map points. Let $\mathcal{S}_{\mathcal{Q}}\in \mathcal{Q}^{\textrm{raw}}$ be the sampled subset of map points\revision{. T}he spatial normalization scale is computed as the median nearest-neighbour distance among sampled map points, while the curvature normalization scale is computed as the median magnitude of the sampled curvature values 
\begin{equation}
\begin{split}
\eta_d & = \operatorname*{median}_{\mathbf{q}\in\mathcal{S}_{\mathcal{Q}}}\left(\min_{\mathbf{q}'\in\mathcal{K}\left(\mathbf{q}\right)} \lVert \mathbf{q} - \mathbf{q}' \rVert\right), \\
\eta_\kappa
    &= \operatorname*{median}_{\mathbf{q}\in\mathcal{S}_{\mathcal{Q}}} \left(|\kappa_q|\right).
\end{split}
\end{equation}
This adaptive normalization finds the typical distance between map points and the typical submap curvature value to ensure stable correspondence selection across scenes with varying point densities and surface characteristics. \revision{Since $\eta_d$ and $\eta_\kappa$ adapt to the local point density and curvature distribution, $\beta$ only balances the two normalized terms rather than compensating for scene-specific scale differences. The supplementary material\textsuperscript{\ref{fn:supplementary}} empirically shows that correspondence quality and pose correction are not strongly sensitive to $\beta$, supporting a fixed value across environments.}

The use of a joint spatial–curvature correspondence score mitigates two key limitations of purely distance-based matching. In geometrically degenerate regions, such as large planar or weakly structured areas, multiple candidate map points often lie at similar Euclidean distances from a query point, leading to ambiguous or unstable correspondences when spatial proximity alone is used. Curvature information captures local surface geometry and helps distinguish structurally informative features, such as edges and corners, from less informative planar regions. However, relying solely on curvature may produce geometrically inconsistent matches. By jointly considering spatial proximity and geometric similarity, the proposed correspondence score \revision{is designed to yield} more reliable and stable data association across diverse environments. The normalization step further prevents high-density or low-information regions from dominating the correspondence search, improving robustness across scenes with varying point densities and surface characteristics.

After selecting the best candidate match for each scan point, correspondences are further filtered using geometric consistency checks to remove unreliable matches. First, a spatial gating step discards pairs whose Euclidean separation exceeds a predefined threshold, ensuring that only locally consistent associations are retained. Once a reasonable initial alignment has been established, an additional filtering step removes correspondences with large point-to-plane residuals relative to the target surface normal, rejecting associations that violate local planar geometry. These filtering steps improve robustness by eliminating outliers before the subsequent optimization stage.

\subsubsection{\revision{Adaptive Registration Formulation}}
After curvature-based data association and correspondence filtering, we obtain the live lidar point cloud, $\mathcal{P}=\left\{\mathbf{p}_i, \revision{\kappa_{p_i}} \right\}$, and the corresponding local map points $\mathcal{Q}=\left\{\mathbf{q}_i, \mathbf{n}_i, \revision{\kappa_{q_i}} \right\}$, where $\mathbf{n}_i\in\mathbb{R}^3$ \revision{denotes the unit normal vector of the map point, and $\kappa_{p_i}$ and $\kappa_{q_i}$ denote the Gaussian curvatures associated with the live and map points, respectively.} \revision{Inspired by the hybrid registration formulation of Lee~\textit{et al.}~\cite{lee2024genz}, we leverage the Gaussian curvature information of the matched map point to determine the ICP residual for each correspondence. When the Gaussian curvatures of both the live scan point and the matched map point exceed a predefined threshold, the point-to-point residual is employed, as both points are likely to correspond to distinctive non-planar geometric structures. The point-to-point ICP residual is defined as
\begin{equation}
\label{eq:p2point_residual}
    \mathbf{r}_i^{\text{p2p}}(\mathbf{T}_{m,s}) = \mathbf{T}_{m,s}\mathbf{p}_i - \mathbf{q}_i,
\end{equation}
where $\mathbf{T}_{m,s}= \left\{\mathbf{R}_{m,s},\mathbf{t}^{s,m}_m\right\}\in SE(3)$ denotes the transformation from the lidar sensor frame to the local submap frame. An
isotropic noise $\mathbf{\Sigma}_{\text{p2p},i}=\sigma_{\text{p2p}}^2\mathbf{1}_3$ is applied to model the point-to-point residual uncertainty. When the Gaussian curvatures of both points are below the threshold, the point-to-plane residual is employed to better exploit the locally planar surface constraints. The corresponding point-to-plane residual is given by}
\begin{equation}
\label{eq:p2plane_residual}
    r_i^{\revision{\text{p2pl}}}(\mathbf{T}_{m,s}) = \mathbf{n}_i^{\top} \mathbf{D} (\mathbf{T}_{m,s}\mathbf{p}_i - \mathbf{q}_i),
\end{equation}
where $\mathbf{D}$ is a constant projection that removes the homogeneous element. 

\revision{To model the uncertainty of the point-to-plane registration residual, we adopt a first-order uncertainty propagation framework similar to~\cite{liu2021balm}, accounting for noise in the scan point, map point, and surface normal. Specifically, the scan-point covariance $\mathbf{\Sigma}_{p_i}$ is obtained by propagating the lidar range-bearing measurement uncertainty to Cartesian coordinates, while the map-point covariance is approximated by an isotropic model $\mathbf{\Sigma}_{q_i}=\sigma_m^2\mathbf{1}_3$. The uncertainty of the estimated surface normal is assumed to lie in the tangent plane and is modelled as $\mathbf{\Sigma}_{n_i}=\sigma_n^2(\mathbf{1}_3-\mathbf{n}_i\mathbf{n}_i^\top)$. Given the point-to-plane residual expressed in~\eqref{eq:p2plane_residual}, the corresponding Jacobians for lidar point $\mathbf{p}_i$, map point $\mathbf{q}_i$, and the normal vector $\mathbf{n}_i$ are
\begin{equation}
\begin{aligned}
    \mathbf{J}_{p_i} &= \mathbf{n}_i^\top\mathbf{R}_{m,s}, ~~~~
    \mathbf{J}_{q_i} = -\mathbf{n}_i^\top, \\
    \mathbf{J}_{n_i} &= \bigl(\mathbf{D}(\mathbf{T}_{m,s}\mathbf{p}_i'-\mathbf{q}_i')\bigr)^\top .
\end{aligned}
\end{equation}
Assuming the three error sources are mutually independent, the residual variance is obtained through first-order uncertainty propagation,
\begin{equation}
    \sigma_{\text{p2pl},i}^2 =
    \mathbf{J}_{p_i}\mathbf{\Sigma}_{p_i}\mathbf{J}_{p_i}^\top +
    \mathbf{J}_{q_i}\mathbf{\Sigma}_{q_i}\mathbf{J}_{q_i}^\top +
    \mathbf{J}_{n_i}\mathbf{\Sigma}_{n_i}\mathbf{J}_{n_i}^\top .
\end{equation}
The detailed derivation of the lidar point-to-plane covariance is provided in the supplementary material\textsuperscript{\ref{fn:supplementary}}.}

\revision{This adaptive residual selection allows the registration formulation to better leverage the local surface geometry constraints, which improves the registration robustness. We stack the selected residuals into a single residual vector 
\begin{equation}
\label{cha4:residual_vector}
    \mathbf{r}(\mathbf{T}_{m,s}) = \begin{bmatrix}
        \mathbf{r}^{\text{p2p}}(\mathbf{T}_{m,s}) \\ \mathbf{r}^{\text{p2pl}}(\mathbf{T}_{m,s})
    \end{bmatrix},
\end{equation}
where $\mathbf{r}^{\text{p2p}}(\mathbf{T}_{m,s})$ and $\mathbf{r}^{\text{p2pl}}(\mathbf{T}_{m,s})$ stack all point-to-point and point-to-plane residuals, respectively. With $N$ point-to-point residuals and $M$ point-to-plane residuals, the corresponding information matrix is 
\begin{equation}
    \boldsymbol{\mathcal{I}} = \begin{bmatrix}
        \bm{\Sigma}_{\text{p2p}}^{-1}  & \mathbf{0} \\
        \mathbf{0}  & \bm{\Sigma}_{\text{p2pl}}^{-1}
    \end{bmatrix}
\end{equation}
where
\begin{equation}
    \begin{split}
        \bm{\Sigma}_{\text{p2p}}  &= \diag\left(\bm{\Sigma}_{\text{p2p},1}, \cdots, \bm{\Sigma}_{\text{p2p},N}\right),  \\ 
        \bm{\Sigma}_{\text{p2pl}} &= \diag\left(\sigma^2_{\text{p2pl},1}, \cdots, \sigma^2_{\text{p2pl},M}\right).
    \end{split}
\end{equation}
The scan-to-map registration is formulated as the following weighted least-squares optimization:}
\begin{equation}
\label{cha4:optimization}
\mathbf{T}^{\star}_{m,s} = \argmin_{\mathbf{T}_{m,s}}~\mathbf{r}(\mathbf{T}_{m,s})^\top~\boldsymbol{\mathcal{I}}~\mathbf{r}(\mathbf{T}_{m,s}).
\end{equation} 

\subsubsection{Degeneracy-Aware Optimization} 

\revision{
While the adaptive registration formulation improves robustness, it may still suffer from weakly constrained directions in extremely degenerate environments dominated by self-similar planar features. We therefore solve the weighted least-squares problem in~\eqref{cha4:optimization} using a degeneracy-aware optimization.} We perturb the pose with a Lie algebra perturbation $\bm{\xi} = [\bm{t}^\top ~~ \bm{\theta}^\top]^\top\in \mathbb{R}^6$, \revision{where $\bm{t}$ and $\bm{\theta}$ denote the translational and rotational components, respectively.} The perturbation $\bm{\xi}$ is applied to an initial guess transformation $\bar{\mathbf{T}}_{m,s}$ as
\begin{equation}
    \mathbf{T}_{m,s} = \exp{(\bm{\xi}^{\wedge})}\bar{\mathbf{T}}_{m,s}.
\end{equation}
Linearizing the residual in \eqref{cha4:residual_vector} and setting the gradient to zero yields the Gauss-Newton normal equation: 
\begin{equation}
\label{cha4:gn_normal_equ}
    \mathbf{H} \bm{\xi}^{\star} = \mathbf{b},
\end{equation}
where the Hessian matrix $\mathbf{H}$ and right-hand-side vector $\mathbf{b}$ are defined as \begin{equation}
    \mathbf{H} = \mathbf{J}^{\top}\boldsymbol{\mathcal{I}}\mathbf{J}, ~~~~~ \mathbf{b} = -\mathbf{J}^\top\boldsymbol{\mathcal{I}}\mathbf{r},
\end{equation}
and where $\mathbf{J}$ is the Jacobian of the residual vector. Readers are referred to~\cite{pomerleau2015review} for more details and derivations about \revision{point cloud registration algorithms.}

The general principle of optimization degeneracy detection relies on checking the eigenvalues of the Hessian matrix~\cite{zhang2016degeneracy, hinduja2019degeneracy}; however, in \revision{ICP-based point cloud registration}, the rotational and translational components operate on different numerical scales. Directly applying eigenvalue-based tests to the original Hessian matrix often leads to unreliable detection of any truly under-constrained directions~\cite{tuna2023x}. \revision{DCReg~\cite{hu2025dcreg} provides a detailed theoretical analysis of this scale imbalance and addresses it by decoupling the translational and rotational subspaces for degeneracy analysis, followed by structure-aware preconditioning of the original coupled optimization problem. Motivated by the same scale-imbalance issue, we employ block scaling to perform degeneracy detection and optimization jointly in a unified scaled space, with the solution subsequently mapped back to the original space.}

To evaluate the relative conditioning of the rotational and translational components, we first marginalize each subspace using the Schur complement. With the partitioned Hessian matrix
\begin{equation}
    \mathbf{H} =
    \begin{bmatrix}
        \mathbf{H}_{t t} & \mathbf{H}_{t \theta} \\
        \mathbf{H}_{\theta t} & \mathbf{H}_{\theta \theta}
    \end{bmatrix},
\end{equation}
applying the Schur complement marginalization yields the marginalized rotational and translational information matrices,
\begin{equation}
    \begin{split}
        \mathbf{H}^{\text{marg}}_{t} &= \mathbf{H}_{t t} - \mathbf{H}_{t \theta} \mathbf{H}_{\theta\theta}^{-1} \mathbf{H}_{\theta t} \\
        \mathbf{H}^{\text{marg}}_{\theta} &= \mathbf{H}_{\theta\theta} - \mathbf{H}_{\theta t} \mathbf{H}_{t t}^{-1} \mathbf{H}_{t \theta},
    \end{split}
\end{equation}
each of which captures the information available in its respective subspace while marginalizing out the other. 
To unify the scale of rotational and translational elements, we define a scaling factor, $\ell$, as the ratio of the dominant eigenvalues of $\mathbf{H}^{\text{marg}}_{\theta}$ and $\mathbf{H}^{\text{marg}}_{t}$, 
\begin{equation}
    \ell = \sqrt{\frac{\lambda_{\max}(\mathbf{H}^{\text{marg}}_{\theta})}{\lambda_{\max}(\mathbf{H}^{\text{marg}}_{t})}}.
\end{equation}
Since we design the system for degenerate scenarios, we use the maximal eigenvalues instead of the sum of the eigenvalues to obtain a more stable unit scaling performance. Since real-world scenarios rarely contain directions that are fully degenerate for either rotation or translation, the proposed scaling factor remains well-behaved. The resulting $\ell$ factor normalizes the effective information across the two subspaces and directly informs the block-scaling matrix, $\mathbf{S}$, enabling consistent degeneracy detection and better-conditioned optimization.

Using our scaling factor, $\ell$, we introduce \revision{the} block-diagonal scaling matrix 
\begin{equation}
    \mathbf{S} = \begin{bmatrix}
        \mathbf{1}_3  & \mathbf{0} \\
        \mathbf{0}    & \ell \mathbf{1}_3
    \end{bmatrix},
\end{equation}
and apply it to obtain the scaled perturbation $\tilde{\bm{\xi}} = \mathbf{S} \bm{\xi}$. Substituting this change of variables into the Gauss-Newton normal equation \eqref{cha4:gn_normal_equ} yields the scaled system
\begin{equation}
\label{cha4:scaled_normal}
    \tilde{\mathbf{H}}\tilde{\bm{\xi}} = \tilde{\mathbf{b}},
\end{equation}
with $\tilde{\mathbf{H}} = \mathbf{S}^{-\top}\mathbf{H}\mathbf{S}^{-1}$ and $\tilde{\mathbf{b}} = \mathbf{S}^{-\top}\mathbf{b}$. After solving for the optimal scaled perturbation, $\tilde{\bm{\xi}}^{\star}$, we map it back to the original incremental perturbation $\bm{\xi}^{\star}$:
\begin{equation}
    \bm{\xi} = \mathbf{S}^{-1}\tilde{\bm{\xi}}^{\star}.
\end{equation}
Notably, the two procedures are algebraically identical. The block-scaling mechanism is a change of variables, it does not alter the solution of ICP optimization, but provides a unit-consistent coordinate system in which eigenvalue analysis and degeneracy detection can be performed reliably.

With the scaled and normalized Hessian matrix $\tilde{\mathbf{H}}$, we then apply the eigen-decomposition
\begin{equation}
    \tilde{\mathbf{H}} = \mathbf{V}\mathbf{\Lambda}\mathbf{V}^{\top}, 
\end{equation}
where $\mathbf{V} = [\mathbf{v}_1 ~~ \cdots ~~ \mathbf{v}_6]$ contains the eigenvectors, and $\mathbf{\Lambda} = \diag(\lambda_1, \cdots, \lambda_6)$ stores the corresponding eigenvalues in descending order. We follow the same setup as~\cite{zhang2016degeneracy, hinduja2019degeneracy} and construct the matrices
\begin{equation}
    \begin{split}
        \mathbf{V}_c &= [\mathbf{v}_1 ~~ \cdots ~~ \mathbf{v}_m] \\
        \mathbf{V}_d &= [\mathbf{v}_{m+1} ~~ \cdots ~~ \mathbf{v}_6] \\
        \mathbf{V}_{wc} &= [\mathbf{V}_c ~~ \mathbf{0} ~~ \cdots ~~ \mathbf{0}],
    \end{split}
\end{equation}
where $\mathbf{V}_c$ contains the well-conditioned eigenvector\revision{s} with $\lambda_i > \lambda_{\text{thresh}}$, $\mathbf{V}_d$ stores the degenerate eigenvectors with $\lambda_i \leqslant \lambda_{\text{thresh}}$, and $\mathbf{V}_{wc}$ contains the well-conditioned eigenvectors while zeroing out the degenerate directions. We determine the degenerate eigenvalues by checking the relative eigenvalue ratio. Specifically, an eigenvalue $\lambda_i$ is considered degenerate if its ratio to the largest eigenvalue, $\lambda_{\text{max}}$, exceeds a predefined eigen-ratio, $\gamma$,
\begin{equation}
    \frac{\lambda_{\text{max}}}{\lambda_i} \geqslant \gamma.
\end{equation}
The corresponding eigenvalue threshold is computed by
\begin{equation}
    \lambda_{\text{thresh}} = \frac{\lambda_{\text{max}}}{\gamma}.
\end{equation}

With the degeneracy detection, we apply the same solution-remapping as illustrated in~\cite{hinduja2019degeneracy} and only update the well-conditioned directions,
\begin{equation}
    \bm{\xi}^{\star}_{c}  = \mathbf{V}^{-\top}\mathbf{V}_{wc}^\top\bm{\xi}^{\star}. 
\end{equation}
We then unscale the computed perturbation and compute the updated pose as 
\begin{equation}
    \hat{\mathbf{T}}_{m,s} =  \exp{\left((\mathbf{S}^{-1}\bm{\xi}^{\star}_{c})^{\wedge}\right)}\bar{\mathbf{T}}_{m,s}.
\end{equation}
To compute the covariance associated with the \revision{\ac{CDAICP}} update, we operate on the scaled Hessian $\tilde{\mathbf{H}}$ and the constructed eigenvector matrices, $\mathbf{V}_c$ and $\mathbf{V}_d$. Since we only update the well-conditioned directions, we retain the well-conditioned eigenvalues, $\bm{\Lambda}_c=\diag{(\lambda_i, \cdots, \lambda_m)}$, and zero out the degenerate components, producing the remapped Hessian, $\tilde{\mathbf{H}}_{rm} = \mathbf{V}\diag{(\bm{\Lambda}_c, \mathbf{0})}\mathbf{V}^\top$. To avoid singularity, we add a small regularization $\epsilon \mathbf{1}_6$, yielding the \revision{regularized} remapped Hessian as
\begin{equation}
    \begin{split}
        \tilde{\mathbf{H}}_{r} &= \tilde{\mathbf{H}}_{rm} + \epsilon \mathbf{1}_6  \\
        & \approx \mathbf{V}_c \bm{\Lambda}_c \mathbf{V}_c^{\top} + \mathbf{V}_d (\epsilon \mathbf{1}_6) \mathbf{V}_d^{\top}.
    \end{split}
\end{equation}
As a result, the scaled covariance can be calculated via \revision{the} pseudo-inverse of the \revision{regularized} remapped Hessian:
\begin{equation}
    \tilde{\mathbf{P}} = \tilde{\mathbf{H}}_{r}^{-1} = \mathbf{V}_c \bm{\Lambda}_c^{-1} \mathbf{V}_c^{\top} + \mathbf{V}_d (\epsilon \mathbf{1}_6)^{-1} \mathbf{V}_d^{\top}.
\end{equation}
Finally, we unscale the covariance to map it back to the original perturbation space:
\begin{equation}
    \hat{\mathbf{P}} = \mathbf{S}^{-1} \tilde{\mathbf{P}} \mathbf{S}. 
\end{equation}

After the \revision{\ac{CDAICP}} alignment converges, we fuse its pose estimate with the motion prior from Doppler odometry to refine the pose estimate. We fuse the \revision{\ac{CDAICP}} results with the prior information in a loosely coupled manner. The initial guess for the sensor-to-map transformation is 
\begin{equation}
    \mathbf{T}^{\text{init}}_{m,s} = \left( \mathbf{T}_{s,k} \check{\mathbf{T}}_{k,v} \mathbf{T}_{v,m} \right)^{-1},
\end{equation}
where $\check{\mathbf{T}}_{k,v}$ is from odometry, $\mathbf{T}_{s,k}$ is the known robot-to-sensor extrinsic, and $\mathbf{T}_{v,m}$ is the transform from the reference submap to the current vertex.

To ensure a smooth and robust ICP correction, we wrap the ICP localization error in a robust cost function. We follow the square root formulation in~\cite{dellaert2006square} and define the ICP localization error as
\begin{equation}
    \mathbf{e}_{\text{icp}} \triangleq \hat{\mathbf{P}}^{-\top/2}\ln{\left(\hat{\mathbf{T}}_{m,s}^{-1}\mathbf{T}_{m,s}^{\text{init}}\right)},
\end{equation}
with $\hat{\mathbf{P}}^{-1/2}$ the matrix square root of $\hat{\mathbf{P}}$. The loosely coupled joint optimization problem is formulated as 
\begin{equation}
    \mathcal{J}_{\text{joint}} = \frac{1}{2}\mathbf{e}_{\text{prior}}^{\top}\mathbf{Q}_{\text{prior}}^{-1}\mathbf{e}_{\text{prior}} + \rho{(\mathbf{e}_{\text{icp}})},
\end{equation}
where $\mathbf{Q}_{\text{prior}}$ is the covariance matrix associated with the prior, $\rho{(\cdot)}$ indicates the robust cost function, \revision{and the prior residual is $\mathbf{e}_{\textrm{prior}} = \ln(\check{\mathbf{T}}_{k,v}\mathbf{T}_{k,v}^{-1})^{\vee}$. The prior term regularizes the localization correction, which helps maintain smooth state updates when the ICP constraints become unreliable.} We use the Cauchy robust cost function with the hyperparameter $c=0.1$~\cite{mactavish2015all}, and solve the joint optimization problem to obtain the posterior pose estimate $\hat{\mathbf{T}}_{k,v}$ via Gauss-Newton using the \texttt{STEAM} optimization library~\cite{BarfootSTEAM}.


\subsubsection{Outlier Rejection}
To prevent unreliable localization updates from corrupting the final pose estimate, we perform a simple post-alignment consistency check between the fused \revision{\ac{CDAICP}} pose, $\hat{\mathbf{T}}_{k,v}$, and the incoming odometry estimate, $\check{\mathbf{T}}_{k,v}$. We compute the relative transform
\begin{equation}
    \mathbf{T}_{\text{diff}} = \hat{\mathbf{T}}_{k,v} \check{\mathbf{T}}_{k,v}^{-1},
\end{equation}
and map it to a 6D vector using the matrix logarithm. The translation and rotation components of this vector quantify how far the joint solution has moved from the odometry estimate.

If either component exceeds preset thresholds, the fused result is considered inconsistent, typically due to incorrect surface alignment or insufficient geometric structure, and the system falls back to using the odometry estimate.
\section{Experiments}
\revision{In this section, we provide a comprehensive evaluation of the proposed \ac{TR} system through simulation analysis, closed-loop field experiments with detailed ablation studies, and runtime evaluation. We first present simulation and offline studies that justify the proposed design choices. We then introduce our experimental setup and the \ac{TR} pipeline variants used for the closed-loop ablation study, followed by field test results in feature-rich campus, feature-limited parking lot, and feature-less airport environments. Finally, we report the computational performance of the individual modules in our \ac{TR} navigation system to provide a detailed runtime breakdown.}

\subsection{\revision{Simulation and Offline Evaluation}}\label{subsec:sim_eval}
\subsubsection{Curvature Estimation Sensitivity}
\revision{We first evaluate the sensitivity of the curvature estimation under varying input and parameter settings. The curvature estimate $\kappa(p_i)$ used by the curvature-based preprocessing (Section~\ref{subsec:curv_preprocessing}, Algorithm~\ref{alg:curvature}) depends on both the local point cloud density and the neighbourhood size $k$ used for the $k$-nearest-neighbour search. To characterize this sensitivity, we construct a synthetic scene consisting of a flat ground plane and five non-planar objects (two spherical caps and three paraboloid/saddle patches) with known analytic Gaussian curvature. We then run the production curvature estimation and clustering procedure while independently varying the point density, via random thinning, and the neighbourhood size $k$, measuring the median relative curvature error and the planar/non-planar misclassification rate against groundtruth over $15$ random scene realizations.

Figure~\ref{fig:curv_sensitivity} summarizes the simulation results. Curvature estimation degrades gradually as the point cloud sparsens, remaining below $15\%$ relative error down to a median point spacing of about $0.08~\mathrm{m}$, before failing completely once the spacing reaches about $0.17~\mathrm{m}$, comparable to the scale of the smallest objects in the scene. For reference, raw scans from the Aeva FMCW lidar used in our field experiments have a median nearest-neighbour spacing of only $0.01$--$0.02~\mathrm{m}$, an order of magnitude denser than this failure point, so the sparsity levels at which curvature estimation degrades are beyond what is encountered in practice. With respect to neighbourhood size, both metrics are poor for small $k$ ($k\!\le\!15$), where the covariance and quadratic fit are underdetermined, and improve sharply by $k\!=\!20$--$30$. Beyond this, the misclassification rate decreases while curvature error slowly rises again due to oversmoothing. The production default $k=30$ lies in the stable region of both metrics.

\begin{figure}[t]
    \centering
    \includegraphics[width=\linewidth]{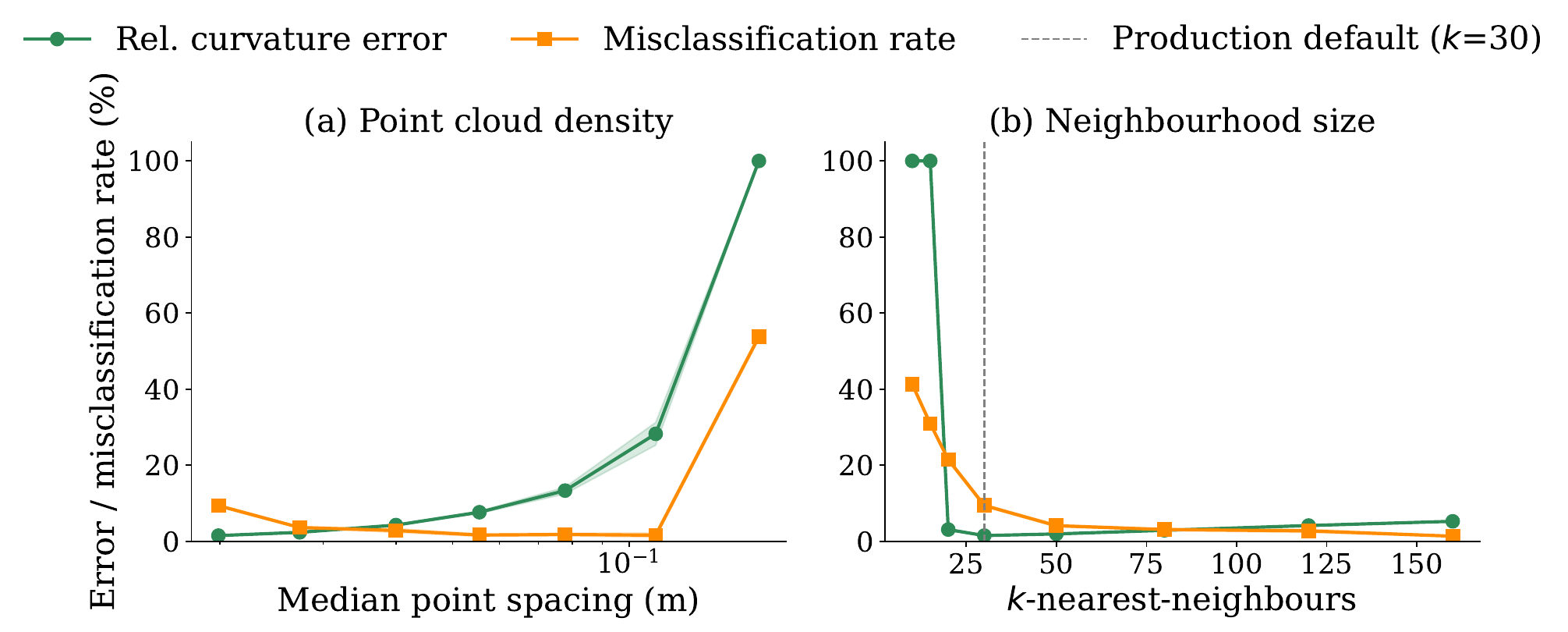}
    \caption{\revision{Sensitivity of curvature estimation to (a) point cloud density, reported as the median nearest-neighbour spacing, and (b) the neighbourhood size $k$, evaluated on a synthetic scene with known groundtruth curvature. Shaded bands denote $\pm 1$ standard deviation over $15$ random scene realizations.}}
    \label{fig:curv_sensitivity}
\end{figure}}

\revision{
\begin{figure}[b!]
    \centering
    \begin{tikzpicture}
    \node[inner sep=0pt, opacity=0.9] (anchor) at (0.0,0.0)
    {\includegraphics[width=.45\textwidth]{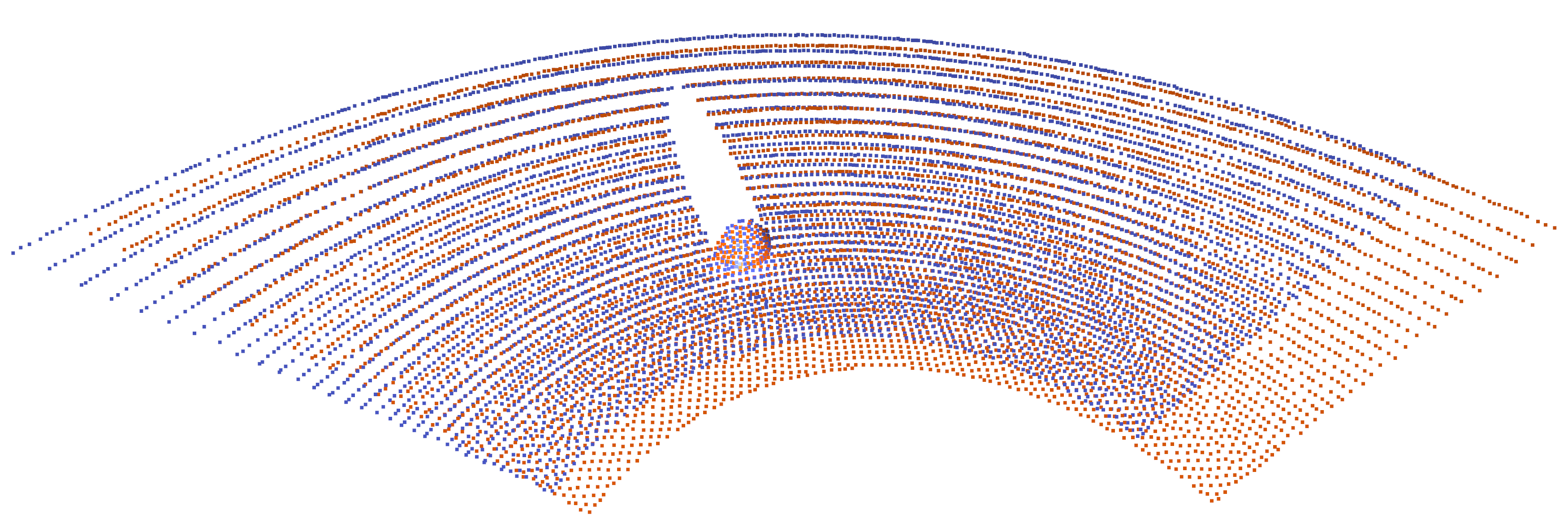}};
    \definecolor{amaranth}{rgb}{0.9, 0.17, 0.31}
    \definecolor{Blue}{rgb}{0.01, 0.28, 1.0}
    \draw[amaranth, thick, dash pattern=on 1.5pt off 1.5pt] (2.5,0.7) -- (2.95,1.1);
    \node[text width=3cm, text=amaranth] at (3.26, 1.3) {\small Target Point Cloud};
    \draw[Blue, thick, dash pattern=on 1.5pt off 1.5pt] (-2.65,1.5) -- (-2.0, 1.0);
    \node[text width=3cm, text=Blue] at (-2.45, 1.73) {\small Source Point Cloud};
    \end{tikzpicture}
    \caption{\revision{One representative degenerate scene used in the simulation study. The environment is dominated by planar surfaces with only sparse non-planar geometry, resulting in limited geometric constraints for registration.}}
    \label{fig:sim_placeholder}
\end{figure}

\subsubsection{Localization Module Ablation Study} We next evaluate how the individual components of the proposed CDA-ICP localization method contribute to registration performance in a representative geometrically degenerate scenario. As illustrated in Figure~\ref{fig:sim_placeholder}, the simulated environment consists predominantly of planar surfaces with only sparse non-planar geometric features, resulting in weak geometric constraints along the degenerate directions. A known transformation is applied to generate the source and target point clouds, providing groundtruth relative pose for quantitative evaluation. All methods are initialized with the same pose perturbation of $\left[0.25, -0.25, 0.25\right]$ metres in translation and $\left[0.0, 0.0, -2.0\right]$ degrees in rotation.

\newcommand{\yes}{\ding{51}}
\newcommand{\na}{--}
\begin{table}[tb!]
    \renewcommand{\arraystretch}{1.3}
    \footnotesize
    \centering
    \caption{\revision{Definition of the five ICP variants (L1--L5) used in the simulated ablation study. Starting from L1, each subsequent variant adds one or more proposed components up to L4, the proposed method; L5 instead replaces L4's solution remapping with a constraint-based optimization alternative. \yes/\na~ indicate whether the column's component is applied.}}
    \label{tab:sim_variant}
    \setlength{\tabcolsep}{3pt}
    \resizebox{\columnwidth}{!}{%
    \begin{tabular}{@{}ccccc@{}}
        \toprule
        \textbf{Variant} & \textbf{Data Association} & \textbf{Residual} & \textbf{Scaling} & \textbf{Degeneracy Handling} \\
        \midrule
        L1 & Standard        & Point-to-Plane & \na  & \na       \\
        L2 & Curvature-enhanced & Point-to-Plane & \na  & Solution Remapping \\
        L3 & Curvature-enhanced & Point-to-Plane & \yes & Solution Remapping \\
        L4 & Curvature-enhanced & Adaptive       & \yes & Solution Remapping \\
        L5 & Curvature-enhanced & Adaptive       & \yes & Constrained Optimization        \\
        \bottomrule
    \end{tabular}%
    }
\end{table}

We consider the five ICP localization variants summarized in Table~\ref{tab:sim_variant} for comparison. L1 uses conventional point-to-plane ICP and serves as the baseline. L2 introduces curvature-enhanced data association and solution remapping for degeneracy handling. L3 further applies the block-diagonal scaling. L4 incorporates the curvature-based adaptive point-to-point/point-to-plane residual formulation, representing the proposed CDA-ICP method. Finally, L5 replaces solution remapping in L4 with a constraint-based optimization approach adopted from~\cite{tuna2025informed}, while retaining the same data association and residual formulation. We select the inequality constraint formulation as a representative constraint-based degeneracy handling strategy due to its relaxed constraints, which potentially provide a larger convergence basin~\cite{tuna2025informed}. Specifically, we define the constraint bounds in the pose perturbation space as $\bm{\varepsilon} = [\delta\bm{t}^\top ~~ \delta\bm{\theta}^\top]^\top \in \mathbb{R}^6$. The constraints are then normalized using the same block-diagonal scaling matrix $\mathbf{S}$ and projected onto the detected degenerate eigenspace $\mathbf{V}_d$, yielding bounded inequality constraints along the degenerate directions. Following the Gauss-Newton linearization of the weighted least-squares problem~\eqref{cha4:optimization} and the scaled normal equation~\eqref{cha4:scaled_normal}, the resulting constrained optimization can be reformulated as the following quadratic program:
\begin{equation}
\label{eq:qp_math}
\begin{split}
    & \min_{\tilde{\bm{\xi}}} \frac{1}{2} \tilde{\bm{\xi}}^\top \mathbf{F}\tilde{\bm{\xi}} + \mathbf{f}^\top\tilde{\bm{\xi}} \\
    & \textrm{s.t.} -|\mathbf{V}_{d}^\top\mathbf{S}\bm{\varepsilon}| \le \mathbf{V}_{d}^\top \tilde{\bm{\xi}} \le |\mathbf{V}_{d}^\top\mathbf{S}\bm{\varepsilon}|,
\end{split}
\end{equation}
where $\mathbf{F} = \tilde{\mathbf{H}}$ and $\mathbf{f} = -\tilde{\mathbf{b}}$. The resulting quadratic program is solved using the QRQP solver~\cite{Andersson2018} provided by CasADi. 


\begin{table}[b!]
    \renewcommand{\arraystretch}{1.3}
    \footnotesize
    \centering
    \caption{\revision{Registration results of the five ICP variants (L1–L5) in the simulated degenerate scene, with the initial translational and rotational error provided for reference. The lowest and second-lowest errors are highlighted in blue and teal, respectively.}}
    \label{tab:sim_icp_results}
    \setlength{\tabcolsep}{3pt}
    \begin{tabular*}{\columnwidth}{@{\extracolsep{\fill}}lcccccc@{}}
        \toprule
        & \textbf{Init.} & \textbf{L1} & \textbf{L2} & \textbf{L3} & \textbf{L4} & \textbf{L5} \\
        \midrule
        Trans.\ Error (m)   & 0.410 & 1.508 & 0.421 & 0.325 & \textcolor{teal}{\textbf{0.164}} & \textcolor{blue}{\textbf{0.116}} \\
        Rot.\ Error (deg)   & 2.000 & 9.337 & 2.471 & 1.998 & \textcolor{teal}{\textbf{1.255}} & \textcolor{blue}{\textbf{0.944}} \\
        \bottomrule
    \end{tabular*}
\end{table}
We summarize the translational and rotational errors with respect to the groundtruth of all five ICP variants in Table~\ref{tab:sim_icp_results}. Conventional point-to-plane ICP (L1) fails to recover the relative pose under the limited geometric constraints, resulting in substantial translation and rotation errors. Introducing curvature-based data association and solution remapping (L2) prevents registration divergence, while incorporating block scaling (L3) improves the registration accuracy. The curvature-enhanced adaptive residual formulation (L4) further reduces the registration errors. The constrained optimization variant L5 achieves slightly better performance compared to the proposed method L4 in this controlled scenario, demonstrating the effectiveness of both degeneracy-handling strategies. Their robustness in closed-loop navigation is subsequently evaluated through comprehensive field experiments in Section~\ref{sec:close-loop-results}.

\begin{table}[b!]
    \renewcommand{\arraystretch}{1.3}
    \footnotesize
    \centering
    \caption{\revision{Eigenvalue ratios of the Hessian matrix at convergence for L2, L3, and L4 in the simulated degenerate scene.}}
    \label{tab:sim_eigen_ratio}
    \setlength{\tabcolsep}{3pt}
    \begin{tabular*}{\columnwidth}{@{\extracolsep{\fill}}lcccccc@{}}
        \toprule
        & $\lambda_{\textrm{max}} \slash \lambda_1$ & $\lambda_{\textrm{max}} \slash \lambda_2$ & $\lambda_{\textrm{max}} \slash \lambda_3$ & $\lambda_{\textrm{max}} \slash \lambda_4$ & $\lambda_{\textrm{max}} \slash \lambda_5$ & $\lambda_{\textrm{max}} \slash \lambda_6$ \\
        \midrule
        \textbf{L2}    & 1.00 & 2.27 & 320.29  & 646.63  & $5.44\times 10^4$ & $9.40\times10^6$ \\
        \textbf{L3}    & 1.00 & 9.73 & 41.94 & 226.67  & 1137.31  & $6.89\times10^5$ \\
        \textbf{L4}    & 1.00 & 9.24 & 29.78 & 36.83 & 119.77 & $8.93\times10^4$ \\
        \bottomrule
    \end{tabular*}
\end{table}
We further investigate the effects of block scaling and adaptive residual selection on the numerical conditioning of the registration problem. Table~\ref{tab:sim_eigen_ratio} reports the eigenvalue ratios $\lambda_{\textrm{max}} \slash \lambda_i$ of the normal matrix at convergence for L2, L3, and L4. Comparing L2 and L3 shows that block scaling substantially reduces the large eigenvalue ratios associated with the weakly constrained directions. The ratios are further reduced in L4 with the curvature-guided adaptive residual formulation, consistent with the conditioning benefits of hybrid registration reported in~\cite{lee2024genz}.

\subsubsection{Convergence and Degeneracy Threshold Sensitivity}
Next, we evaluate the convergence robustness of CDA-ICP under varying initial pose errors and degeneracy thresholds. We use the same simulated degenerate scene and jointly vary the initial-pose perturbation and the degeneracy threshold $\gamma$. Five levels of initial-pose perturbation are considered for the convergence analysis, with the translational and rotational magnitudes initialized at $\left(0.075~\textrm{m}, 0.375^{\circ}\right)$ and doubled at each successive level, reaching $\left(1.2~\textrm{m}, 6.0^{\circ}\right)$ at the largest perturbation (level 5). For each level, we sample $30$ random perturbations and use the same initial perturbations to evaluate registration convergence across all values of the degeneracy threshold $\gamma$. We evaluate the sensitivity to the degeneracy threshold parameter $\gamma$ using $18$ threshold values selected based on the eigen-ratios of this simulated scene. These values are chosen to cover cases with different numbers of detected degenerate directions, denoted as $n_{\textrm{deg}}$, allowing us to assess the convergence behaviour across different degeneracy classifications. To compute the convergence rate, we consider one registration trial converged when the final pose errors fall within into $\|\Delta t\|<0.2$ m and $|\Delta \theta| < 2^{\circ}$.

We summarize the results of the convergence analysis across the initial-pose perturbation and degeneracy thresholds in Figure~\ref{fig:sim_convergence}. In Figure~\ref{fig:sim_convergence}a, we visualize the convergence rate for each perturbation level and degeneracy threshold $\gamma$, computed over the $30$ randomly sampled initial perturbations, into a heatmap. Figure~\ref{fig:sim_convergence}b summarizes the results across all perturbation levels and presents the translational and rotational errors together with the number of detected degenerate directions $n_{\textrm{deg}}$ for each $\gamma$. The results demonstrate a broad convergence basin for the proposed CDA-ICP. For the perturbation levels 1--3, high convergence rates are maintained over a wide range of degeneracy thresholds, approximately from $200$ to $7\times10^4$. Perturbation level 5 represents a particularly challenging scenario as the initial perturbations approach or exceed the nearest-neighbour correspondence threshold, leading to invalid data association. Despite these large initial perturbations, the proposed method remains robust to the choice of $\gamma$ over $200$ to $1000$, where one direction is consistently detected as degenerate ($n_{\textrm{deg}}=1$). In summary, these results indicate that the proposed CDA-ICP maintains a large convergence basin and is not sensitive to a narrowly tuned degeneracy threshold.  

\begin{figure}[!b]
    \centering
    \begin{subfigure}[t]{\linewidth}
        \centering
        \includegraphics[width=\linewidth]{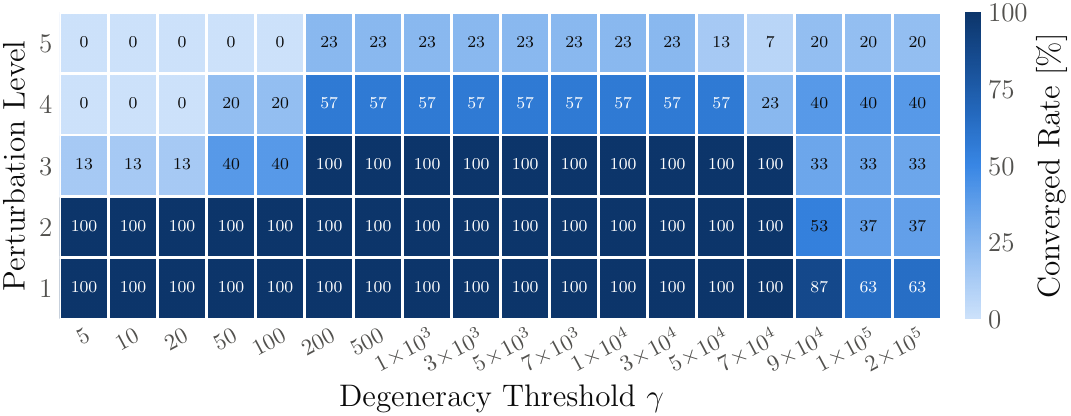 }
        \caption{\revision{Convergence analysis}}
    \end{subfigure}
    \begin{subfigure}[t]{\linewidth}
        \centering
        \includegraphics[width=\linewidth]{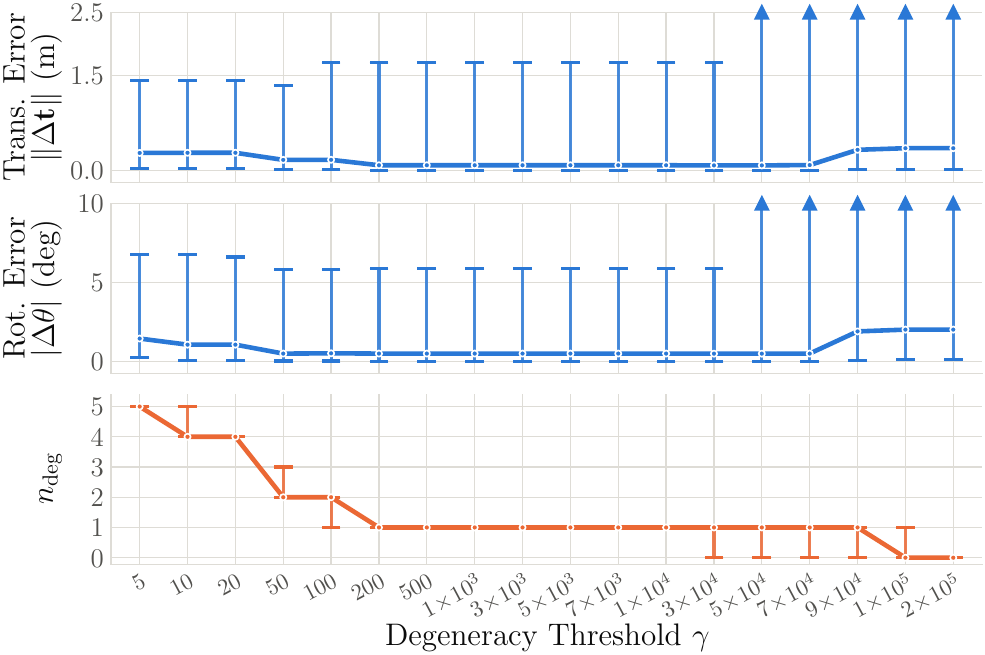}
        \caption{\revision{Degeneracy threshold sensitivity}}
    \end{subfigure}
    \caption{\revision{Convergence and degeneracy threshold sensitivity analysis. (a) Convergence rate over different initial-pose perturbation levels and degeneracy thresholds $\gamma$. (b) Translational and rotational errors and the number of detected degenerate directions $n_{\textrm{deg}}$ across all perturbation levels for each $\gamma$. The bars indicate the minimum and maximum values, and upward arrows denote errors exceeding the displayed $y$-axis limits.}}
    \label{fig:sim_convergence}
\end{figure}

\subsubsection{Odometry Covariance Impact Analysis}
Finally, we evaluate the effect of propagating the estimated odometry covariance to the localization module using recorded data from the field-test environments introduced in Section~\ref{sec:exp_setup_env}. To capture different levels of geometric constraint, we select representative sequences from the feature-less, feature-limited, and feature-rich environments. For each sequence, we replay the same recorded data for both teach and repeat such that differences between the trajectories primarily reflect estimation inconsistency. We compare the proposed odometry covariance against five fixed covariance settings (S1--S5),
\begin{equation}
    \bm{\Sigma}_{\textrm{odom}} = s\times\textrm{diag}\left(10^{-3}, 10^{-3}, 10^{-3}, 10^{-6}, 10^{-6}, 10^{-6}\right) 
\end{equation}
with $s\in\left\{0.01, 0.1, 1, 10, 100\right\}$, while keeping all other system parameters unchanged.

Table~\ref{tab:cov_ablation_results} summarizes the resulting lateral RMSE. In the feature-less environment, the system benefits from using the estimated odometry covariance, reducing the RMSE compared with all fixed covariance settings. In the feature-limited and feature-rich environments, frequent successful localization updates continuously correct the state uncertainty, making the effect of the propagated odometry covariance less pronounced. In contrast, localization updates provide limited corrections in the feature-less environment, where the improved performance with the estimated covariance suggests a potential benefit from more appropriate uncertainty weighting. Overall, these results demonstrate the benefit of the proposed covariance estimation, particularly when reliable geometric constraints are limited.
\begin{table}[t!]
    \renewcommand{\arraystretch}{1.3}
    \footnotesize
    \centering
    \caption{\revision{Odometry covariance ablation study. Lateral RMSE for the proposed estimated odometry covariance (Ours) versus five fixed covariance settings (S1--S5), each a diagonal covariance scaled by a constant factor $s\in\{0.01, 0.1, 1, 10, 100\}$ respectively, evaluated on representative feature-less, feature-limited, and feature-rich sequences. The lowest RMSE for each environment is highlighted in bold.}}
    \label{tab:cov_ablation_results}
    \setlength{\tabcolsep}{3pt}
    \begin{tabular*}{\columnwidth}{@{\extracolsep{\fill}}lcccccc@{}}
        \toprule
        \textbf{RMSE (m)} & \textbf{Ours} & \textbf{S1} & \textbf{S2} & \textbf{S3} & \textbf{S4} & \textbf{S5} \\
        \midrule
        Feature-Less    & \textbf{0.180} & 0.191 & 0.190 & 0.201 & 0.295 & 0.304 \\
        Feature-Limited & 0.108 & 0.110 & \textbf{0.104} & 0.113 & 0.144 & 0.134 \\
        Feature-Rich    & 0.142 & 0.143 & 0.167 & \textbf{0.125} & 0.143 & 0.137 \\
        \bottomrule
    \end{tabular*}
\end{table}
}

\subsection{Experimental Setup and \revision{\ac{TR} Pipeline Variants}}
\label{sec:exp_setup_env}
We perform all experiments using a Warthog \ac{UGV} equipped with an Aeva Aeries II \ac{FMCW} lidar sensor, set to use a $120^{\circ}$ horizontal field of view, a $19.2^{\circ}$ vertical field of view, and a $40 \ \mathrm{m}$ range. The sensor operates at $10 \ \mathrm{Hz}$ and is equipped with an IMU, from which we use gyroscope measurements. 

We use three environments that differ in geometric structure to compare and evaluate the proposed pipeline in environments ranging from feature-less to feature-rich. Our pipeline is tested on \revision{$1.7 \ \mathrm{km}$} of closed-loop driving, and we perform a comparative study using an additional \revision{$5.3 \ \mathrm{km}$} of closed-loop tests\footnote{\revision{Reported travel distance excludes trials that failed the navigation task.}}. The routes used in our evaluation are:

\begin{itemize}   
    \item Airport Route\revision{s} — \revision{Three routes} collected on an airport runway. The runway is feature-less, therefore rocks are added to the scene to introduce sparse geometric structure.
    
    \item \revision{Parking Lot Route — One route collected in a feature-limited parking lot. Sparse buildings, trees, fake rocks, and terrain variation (hills) provide limited geometric constraints, while the surrounding open paved areas create extended regions with weak geometric structure. }

    \item Campus Route\revision{s} — \revision{Two routes} with many clear geometric features\revision{,} including buildings, cars, trees, and a fence. \revision{These feature-rich routes provide} the baseline against which we compare our pipeline to standard \ac{LTR} performance.
    
\end{itemize}

\begin{figure}[t]
    \centering
    \includegraphics[width=0.95\linewidth]{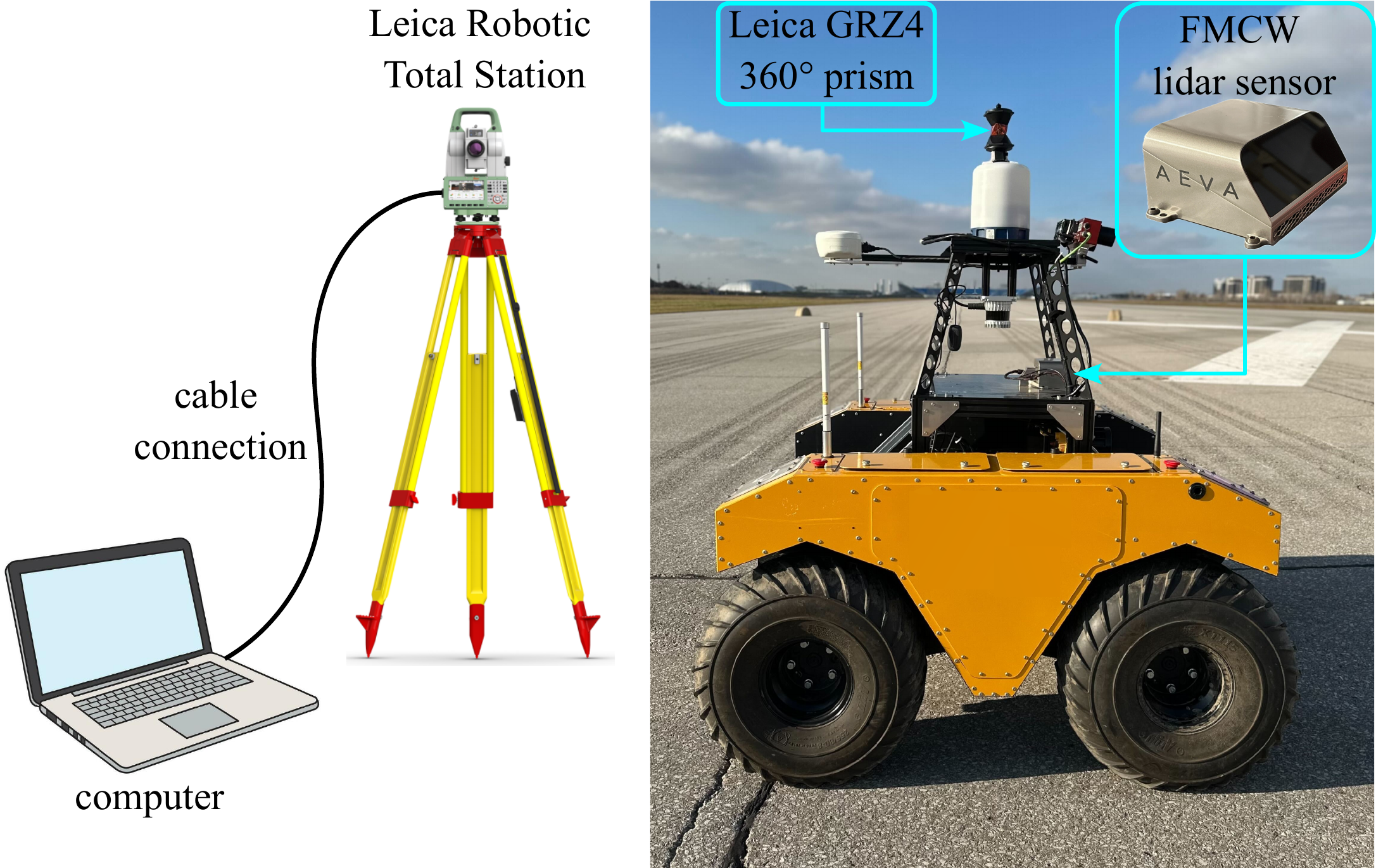}
    \caption{Warthog unmanned ground vehicle (UGV) equipped with an Aeva Aeries II FMCW lidar sensor and a survey-grade prism. A Leica robotic total station tracks the prism to provide millimetre-level groundtruth position measurements for closed-loop navigation field experiments.}
    \label{fig:warthog_exp}
\end{figure}

We obtained the groundtruth measurements on the Campus Route and the Airport Route using a Leica Nova MS50 MultiStation, which continuously tracks a Leica GRZ4 $360^\circ$ prism mounted on a Warthog \ac{UGV}, as seen in Figure~\ref{fig:warthog_exp}. The Leica total station provides millimetre-level positional accuracy under clear line-of-sight conditions, making it well suited for our purpose of evaluating closed-loop performance in feature-less environments.

\begin{table*}[tb]
    \renewcommand{\arraystretch}{1.5}
    \footnotesize
    \centering
    \caption{\revision{Comparison of the six Teach and Repeat (T\&R) pipeline variants evaluated in the closed-loop ablation study. Starting from the default LT\&R pipeline (V1), each subsequent variant progressively introduces one proposed component. The table summarizes the preprocessing, odometry, data association, and localization components, which are the components that differ across the six variants. Newly introduced components are highlighted in different colours to emphasize the incremental progression from V1 to V6.}}
    \label{tab:tr_pipeline_variants}
    \renewcommand{\tabularxcolumn}[1]{m{#1}} 
    \newcolumntype{H}{>{\hsize=0.43\hsize\centering\arraybackslash}X} 
    \newcolumntype{K}{>{\hsize=0.58\hsize\centering\arraybackslash}X} 
    \newcolumntype{L}{>{\hsize=0.79\hsize\centering\arraybackslash}X} 
    \newcolumntype{D}{>{\hsize=0.85\hsize\centering\arraybackslash}X} 
    \newcolumntype{J}{>{\hsize=2.35\hsize\centering\arraybackslash}X} 
    \begin{tabularx}{\linewidth}{H K L L J}
        \toprule
        \textbf{Variant}    & \textbf{Preprocessing} & \textbf{Odometry} & \textbf{Data Association} &  \textbf{Localization}  \\
        \midrule
        V1      &  Uniform         & ICP (point-to-plane)      & Standard & ICP (point-to-plane) \\
        V2      &  Uniform         & \textbf{\color{magenta}{Doppler-Inertial}} & Standard & ICP (point-to-plane) \\
        V3      &  \textbf{\color{orange}{Curvature}} & Doppler-Inertial & Standard & ICP (point-to-plane) \\
        V4      &  Curvature & Doppler-Inertial & \textbf{\color{purple}{Curvature-enhanced}} & \textbf{\color{purple}{ICP (point-to-plane), Block Scaling, Solution Remapping}} \\
        \textbf{V5 (Ours)}     &  Curvature & Doppler-Inertial & Curvature-enhanced & \textbf{\color{blue}{ICP (adaptive), Block Scaling, Solution Remapping}} \\
        V6      &  Curvature & Doppler-Inertial & Curvature-enhanced & \textbf{\color{teal}{ICP (adaptive), Block Scaling, Constrained Optimization}} \\
        \bottomrule
    \end{tabularx}
\end{table*}

\revision{To evaluate the contribution of each component in the proposed \ac{TR} system, we compare the six pipeline variants summarized in Table~\ref{tab:tr_pipeline_variants}. Beginning with the default \ac{LTR} pipeline~\cite{Burnett2022Ready} (V1), we progressively introduce the proposed Doppler-inertial odometry (V2), curvature-based preprocessing (V3), and point-to-plane ICP with curvature-enhanced data association, block scaling and solution remapping (V4). We then enable the adaptive point-to-point/point-to-plane registration formulation (V5), which forms the proposed \ac{TR} pipeline. Finally, we replace the solution-remapping strategy in V5 with the constraint-based optimization approach introduced in~\eqref{eq:qp_math} (V6), while keeping all other pipeline components unchanged. The six pipeline variants are evaluated in the subsequent closed-loop field experiments. By comparing their performance under identical experimental conditions, we isolate the contribution of each proposed component to the overall navigation system.}

\subsection{\revision{Closed-Loop Navigation Results}}
\label{sec:close-loop-results}

\begin{figure*}[!t]
    \centering
    \begin{subfigure}[t]{\textwidth}
        \centering
        \includegraphics[width=0.95\textwidth]{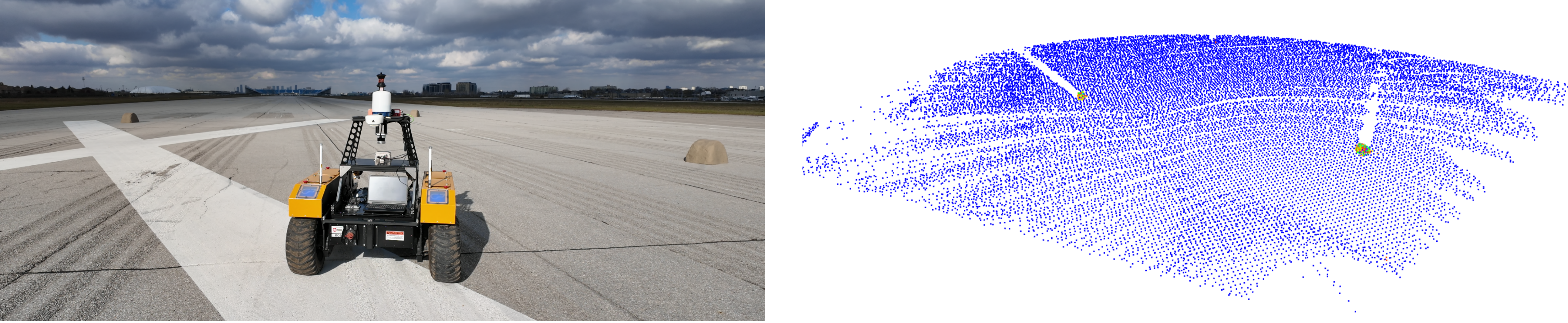}
        \caption{Airport Route \vspace{5pt}}
    \end{subfigure}
    \begin{subfigure}[t]{\textwidth}
        \centering
        \includegraphics[width=0.95\textwidth]{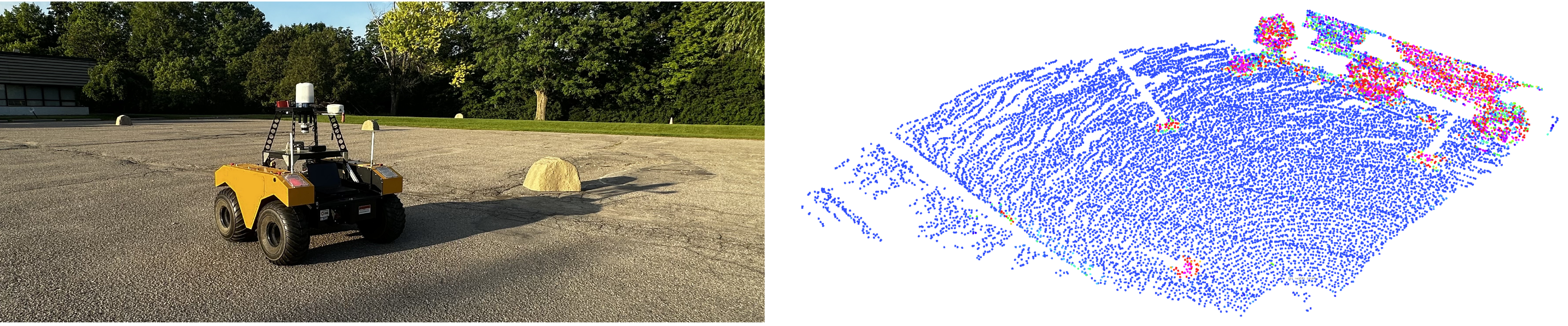}
        \caption{\revision{Parking Lot Route} \vspace{5pt}}
    \end{subfigure}
    \begin{subfigure}[t]{\textwidth}
        \centering
        \includegraphics[width=0.95\textwidth]{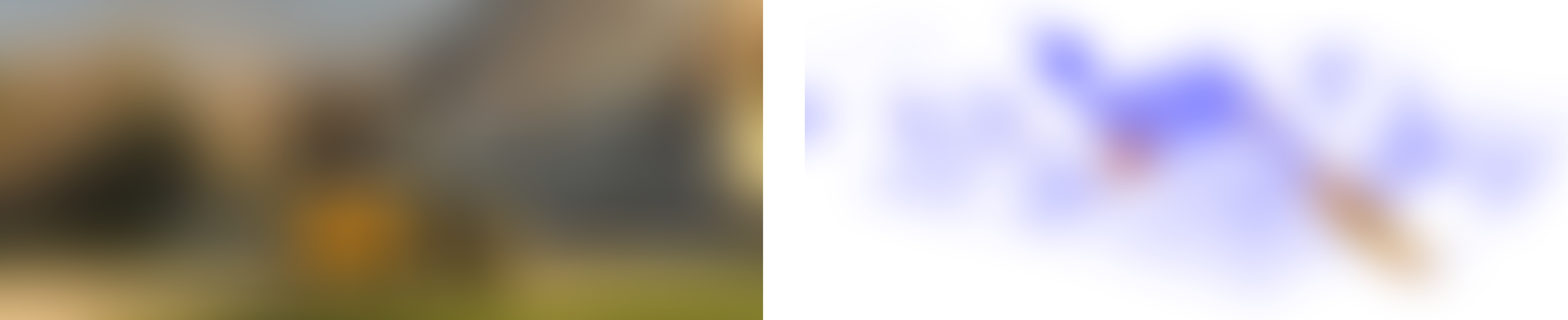}
        \caption{Campus Route}
    \end{subfigure}
    \caption{Experimental environments used for field evaluation. The left column shows representative images captured during the experiments, while the right column presents the corresponding lidar point clouds coloured by curvature, where blue indicates near-zero curvature and non-planar features appear in different colours. The experiments were conducted across environments with varying geometric structure, ranging from feature-rich (campus route\revision{s}), feature-limited (\revision{parking lot} route), to feature-less (airport runway route\revision{s}). Image (c) is intentionally blurred for double-blind review.}
    \label{fig:env_pcloud}
\end{figure*}

\revision{To quantitatively evaluate the proposed \ac{TR} system, we report the lateral \ac{RMSE} and maximum lateral error computed using Leica total station groundtruth measurements.} The lateral \revision{error is computed as} the perpendicular deviation of the repeat trajectory from the reference teach trajectory. The teach and repeat trajectories are aligned by the cumulative arclength along the path. At each point along the path, the vector difference between teach and repeat positions is projected onto the tangent of the reference path, and the component orthogonal to this tangent is the lateral error. The root mean square value of this error over the moving segment constitutes the lateral \ac{RMSE}, \revision{while the maximum lateral error is reported to characterize the worst-case path-tracking deviation.} 

\begin{figure*}[t]
    \centering
    \includegraphics[width=\textwidth]{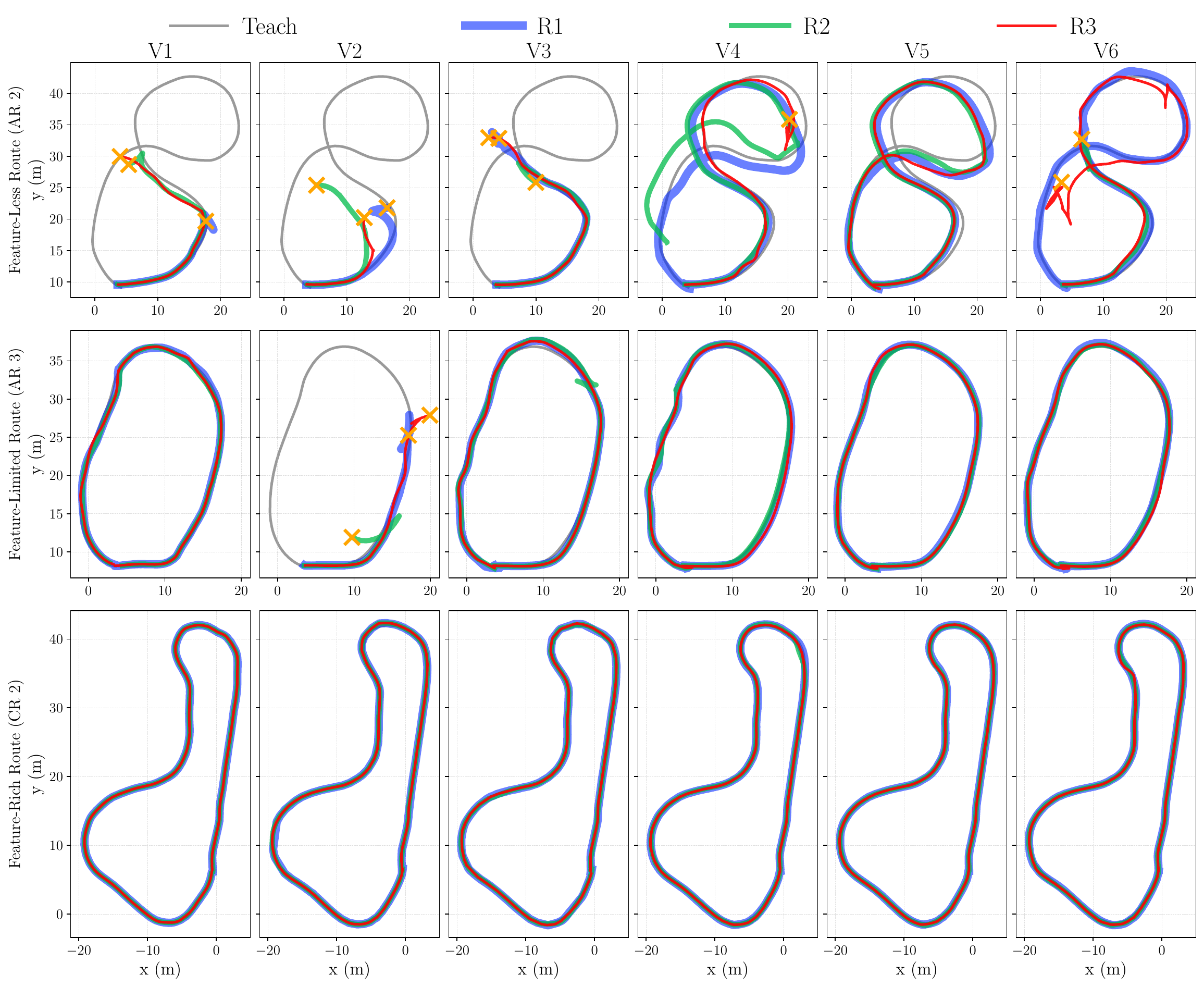}
    \caption{\revision{Closed-loop navigation trajectories (top-down view) of the six T\&R pipeline variants in representative feature-less (AR2), feature-limited (AR3), and feature-rich (CR1) environments. The grey trajectory denotes the teach pass, while the blue, green, and red trajectories correspond to the three repeat trials (R1--R3), evaluated using millimetre-level groundtruth measurements from a Leica total station. The \textcolor{orange}{\bm{$\times$}} symbol indicates that the robot failed the navigation task and required human intervention for safety. The proposed pipeline (V5) is the only variant that consistently completes all closed-loop navigation tasks across environments ranging from feature-rich to feature-less, while maintaining competitive path-tracking accuracy.}}
    \label{fig:field_test_results}
\end{figure*}

\revision{We categorize the six evaluation routes into three levels of geometric complexity. Airport Routes 1 and 2 (AR1 and AR2) are feature-less environments, where we included fake rocks to provide minimal geometric features on a completely flat airport runway. Airport Route 3 (AR3) and the Parking Lot Route (PLR) are feature-limited environments. AR3 is designed such that at least two fake rocks remain visible throughout the route, while PLR contains sparse buildings, trees, fake rocks, and terrain variations (hills), providing moderate geometric constraints. Campus Routes 1 and 2 (CR1 and CR2) represent feature-rich environments with abundant geometric structures, including buildings, trees, parked vehicles, and a fence. The same eigen-ratio threshold ($\gamma=150$) is used for degeneracy detection in all degeneracy-aware pipeline variants (V4--V6) across the six evaluation routes. } \revision{Table~\ref{tab:field_exp_results} summarizes the results over three repeated trials for each route and pipeline variant, where average values are reported when all three trials successfully complete the navigation task. Figure~\ref{fig:field_test_results} illustrates example closed-loop trajectories from the feature-less (AR2), feature-limited (AR3), and feature-rich (CR1) environments to provide further insight into the closed-loop navigation behaviour of the six pipeline variants.}

\newcommand{\failed}{\textcolor{magenta}{\ding{55}}}
\begin{table*}[tb]
    \renewcommand{\arraystretch}{1.2}
    \small
    \centering
    \caption{\revision{Quantitative comparison of the six \ac{TR} pipeline variants in the closed-loop field experiments. AR, PLR, and CR denote the Airport Routes, Parking Lot Route, and Campus Routes, respectively. The six routes are categorized as Feature-Less (AR1–AR2), Feature-Limited (AR3 and PLR), and Feature-Rich (CR1–CR2). R1--R3 correspond to three repeated trials, and \textbf{Ave.} denotes the average over the three trials. Average values are reported when all three trials successfully complete the navigation task. Lateral RMSE and maximum lateral error are computed using Leica total station groundtruth measurements. The lowest and second-lowest average errors for each route are highlighted in blue and teal, respectively.}}
    \label{tab:field_exp_results}
    \setlength{\tabcolsep}{4pt}   
    \resizebox{\textwidth}{!}{%
    \begin{tabular}{@{}cc *{6}{cccc}@{}}
        \toprule
        & & \multicolumn{8}{c}{\textbf{\large{Feature-Less}}} & \multicolumn{8}{c}{\textbf{\large{Feature-Limited}}} & \multicolumn{8}{c}{\textbf{\large{Feature-Rich}}} \\
        \cmidrule(lr){3-10} \cmidrule(lr){11-18} \cmidrule(lr){19-26}
        & & \multicolumn{4}{c}{AR 1} & \multicolumn{4}{c}{AR 2} & \multicolumn{4}{c}{AR 3}
            & \multicolumn{4}{c}{PLR} & \multicolumn{4}{c}{CR 1} & \multicolumn{4}{c}{CR 2} \\
        \cmidrule(lr){3-6} \cmidrule(lr){7-10} \cmidrule(lr){11-14}
        \cmidrule(lr){15-18} \cmidrule(lr){19-22} \cmidrule(lr){23-26}
        & & R1 & R2 & R3 & \textbf{Ave.}\ & R1 & R2 & R3 & \textbf{Ave.}\ & R1 & R2 & R3 & \textbf{Ave.}\
            & R1 & R2 & R3 & \textbf{Ave.}\ & R1 & R2 & R3 & \textbf{Ave.}\ & R1 & R2 & R3 & \textbf{Ave.}\ \\
        \midrule
        \multirow{3.5}{*}{ V1 } & \makecell{Lateral\\ RMSE (m)}
            & \failed & \failed & \failed & NA & \failed & \failed & \failed & NA &  0.207  &  0.264  &  0.206  & 0.226 & \failed & \failed & \failed & NA &  0.060  &  0.062  &  0.064  & \color{blue}{\textbf{0.062}} &  0.067  &  0.068  &  0.066  & \color{blue}{\textbf{0.067}}  \\
        \addlinespace[2pt] \cmidrule(lr){2-2} \addlinespace[2pt]
                              & \makecell{Max Lateral\\ Error (m)}
            & \failed & \failed & \failed & NA & \failed & \failed & \failed & NA &  0.563  &  0.820  &  0.570  & 0.651 & \failed & \failed & \failed & NA &  0.188  &  0.190  &  0.200  & \color{blue}{\textbf{0.193}} &  0.209  &  0.207  &  0.208  &  \color{blue}{\textbf{0.208}} \\
        \midrule
        \multirow{3.5}{*}{ V2 } & \makecell{Lateral\\ RMSE (m)}
            & \failed & \failed & \failed & NA & \failed & \failed & \failed & NA & \failed & \failed & \failed  & NA &  0.053  &  0.055  &  0.059  & \color{teal}{\textbf{0.056}} &  0.103  &  0.111  &  0.113  & 0.109 &  0.072  &  0.071  &  0.069  & \color{teal}{\textbf{0.071}} \\
        \addlinespace[2pt] \cmidrule(lr){2-2} \addlinespace[2pt]
                              & \makecell{Max Lateral\\ Error (m)}
            & \failed & \failed & \failed & NA & \failed & \failed & \failed & NA & \failed & \failed & \failed & NA &  0.193  &  0.196  &  0.183  & 0.191 &  0.416  &  0.460  &  0.435  & 0.437 &  0.334  &  0.323  &  0.324  & 0.327 \\
        \midrule
        \multirow{3.5}{*}{ V3 } & \makecell{Lateral\\ RMSE (m)}
            & \failed & \failed & \failed & NA & \failed & \failed & \failed & NA &  0.293 &  1.594  &  0.302  & 0.730 &  0.055  &  0.058  &  0.053  & \color{blue}{\textbf{0.055}} &  0.100  &  0.102  &  0.098  & \color{teal}{\textbf{0.100}} &  0.085  &  0.081  &  0.075  & 0.080 \\
        \addlinespace[2pt] \cmidrule(lr){2-2} \addlinespace[2pt]
                              & \makecell{Max Lateral\\ Error (m)}
            & \failed & \failed & \failed & NA & \failed & \failed & \failed & NA &  0.710  &  5.121  &  0.836  & 2.222 &  0.207  &  0.167  &  0.162  & \color{blue}{\textbf{0.179}} &  0.297  &  0.259  &  0.313  & \color{teal}{\textbf{0.290}} &  0.398  &  0.337  &  0.302  & 0.346 \\
        \midrule
        \multirow{3.5}{*}{ V4 } & \makecell{Lateral\\ RMSE (m)}
            & \failed & \failed & \failed & NA &  1.234  &  1.977  & \failed & NA &  0.311  &  0.208  &  0.339  & 0.286 &  0.066  &  0.076  &  0.090  & 0.077 &  0.086  &  0.123  &  0.101  & 0.103 &  0.101  &  0.087  &  0.098  & 0.095 \\
        \addlinespace[2pt] \cmidrule(lr){2-2} \addlinespace[2pt]
                              & \makecell{Max Lateral\\ Error (m)}
            & \failed & \failed & \failed & NA &  2.968  &  3.784  & \failed & NA &  0.690  &  0.646  &  0.625  & 0.654 &  0.162  &  0.263  &  0.365  & 0.263 &  0.239  &  0.538  &  0.276  & 0.351 &  0.298  &  0.260  &  0.306  & 0.288 \\
        \midrule
        \multirow{3.5}{*}{ \makecell{\textbf{V5} \\ \textbf{(Ours)}} } & \makecell{Lateral\\ RMSE (m)}
            &  0.292  &  0.241  & 0.525 & \color{blue}{\textbf{0.353}} &  1.401  &  1.553  &  1.541  & \color{blue}{\textbf{1.498}} &  0.117  &  0.105  &  0.126  & \color{blue}{\textbf{0.116}} &  0.067  &  0.077  &  0.073  & 0.072 &  0.095  &  0.099  &  0.104  & 0.099 &  0.099  &  0.069  &  0.108  & 0.092 \\
        \addlinespace[2pt] \cmidrule(lr){2-2} \addlinespace[2pt]
                              & \makecell{Max Lateral\\ Error (m)}
            &  0.681  &  0.637  &  1.310  & \color{blue}{\textbf{0.876}} &  2.946  &  3.657  &  3.297  & \color{blue}{\textbf{3.300}} &  0.300  &  0.242  &  0.451  & \color{blue}{\textbf{0.331}} &  0.263  &  0.304  &  0.198  & 0.255 &  0.297  &  0.294  &  0.322  & 0.304 &  0.256  &  0.218  &  0.281  & 0.252 \\
        \midrule
        \multirow{3.5}{*}{ V6 } & \makecell{Lateral\\ RMSE (m)}
            & \failed & \failed & \failed & NA &  0.318  & \failed & \failed & NA &  0.201  &  0.184  &  0.251  & \color{teal}{\textbf{0.212}} &  0.053  &  0.078  &  0.052  & 0.061 &  0.103  &  0.101  &  0.102  & 0.102 &  0.070  &  0.068  &  0.103  & 0.080 \\
        \addlinespace[2pt] \cmidrule(lr){2-2} \addlinespace[2pt]
                              & \makecell{Max Lateral\\ Error (m)}
            & \failed & \failed & \failed & NA &  0.932  & \failed & \failed & NA &  0.478  &  0.430  &  0.580  & \color{teal}{\textbf{0.496}} &  0.135  &  0.271  &  0.140  & \color{teal}{\textbf{0.182}} &  0.269  &  0.297  &  0.322  & 0.296 &  0.157  &  0.201  &  0.291  & \color{teal}{\textbf{0.216}} \\
        \bottomrule
    \end{tabular}%
    }\par
    \vspace{2pt}
    {\footnotesize \raggedright \failed~denotes a trial that failed the navigation task.\par}
\end{table*}
\paragraph{\revision{Feature-Less Routes (AR1 and AR2)}}
\revision{The results on AR1 and AR2 demonstrate the effectiveness of the proposed \ac{TR} navigation system under severe geometric degeneracy.} \revision{These two routes are} extremely feature-sparse, containing long stretches of asphalt, grass, and occasional rock features \revision{(see Figure~\ref{fig:env_pcloud})}. \revision{Variants without degeneracy-aware localization (V1–V3) consistently fail to complete the navigation task. Our proposed navigation pipeline (V5) is the only variant that consistently completes all trials in both feature-less routes. Although it results in average closed-loop lateral tracking errors of approximately $0.35$ m and $1.50$ m on AR1 and AR2, respectively, we argue that in such severely degenerate environments, successful mission completion should take precedence over precise trajectory tracking. Compared with V4, which employs only point-to-plane ICP residuals, the proposed curvature-enhanced adaptive point-to-point/point-to-plane registration further improves localization robustness by better exploiting the available geometric information during scan-to-map registration. Consequently, while V4 successfully completes two of the three trials in AR2, as illustrated in the top row of Figure~\ref{fig:field_test_results}, it fails in the remaining trial and in all trials on AR1. In contrast, V6 replaces the proposed solution remapping with a representative constraint-based optimization strategy. Although V6 achieves an excellent tracking accuracy in one AR2 trial, with a lateral RMSE of only $0.318$ m (see V6-R1 in the top row of Figure~\ref{fig:field_test_results}), its performance is not consistent across repeated runs. These results suggest that constraint-based optimization has the potential to achieve accurate localization in feature-less environments. However, more systematic algorithm design is required to improve navigation robustness, while broader field experiments are needed to assess its behaviour under closed-loop operation.
}

\paragraph{\revision{Feature-Limited Routes (AR3 and PLR)}}
\revision{The feature-limited environments further illustrate the contribution of the proposed components. The contrasting performance of V1 and V2 demonstrates the role of geometric information in the odometry module. On AR3, the default \ac{LTR} pipeline (V1) successfully completes all trials, whereas the geometry-free Doppler odometry variant (V2) fails. In contrast, V1 fails on PLR while V2 successfully completes all trials. These results suggest that leveraging geometric constraints in the odometry module can be beneficial when reliable geometric features are available, but may also lead to failures as the available features become sparse or unreliable. This observation motivates further investigation into effectively integrating Doppler odometry with geometric constraints while accounting for geometric degeneracy. Comparing V2 and V3 further demonstrates that the proposed curvature-based preprocessing improves localization robustness, enabling successful navigation on AR3 by preserving more geometrically informative points for scan-to-map matching. Finally, our proposed pipeline (V5) achieves the lowest average lateral RMSE of only $0.116$ m on AR3 (shown in the middle row of Figure~\ref{fig:field_test_results}), where geometric constraints remain sparse. In PLR, additional buildings and trees provide stronger localization constraints, resulting in comparable performance among the Doppler odometry-based pipeline variants (V2–V6), with average RMSE values differing by only approximately $2$ cm.
}

\paragraph{\revision{Feature-Rich Routes (CR1 and CR2)}}
\revision{In the feature-rich campus environments, all six pipeline variants successfully complete the navigation task with similar navigation performance, as illustrated by the representative trajectories in the bottom row of Figure~\ref{fig:field_test_results}, where the repeat trajectories almost completely overlap. The default \ac{LTR} pipeline achieves the lowest lateral RMSE of approximately $6$ cm by fully exploiting the abundant geometric constraints available for both odometry and localization. The Doppler odometry-based pipeline variants (V2--V6) achieve slightly higher tracking errors but comparable closed-loop navigation performance while benefiting from a significantly lower computational cost, as discussed in Section~\ref{subsec:runtime}. Among these variants, the average lateral RMSE differs by only $2$--$3$ cm, with the proposed pipeline (V5) achieving average errors of $0.099$ m and $0.092$ m on CR1 and CR2, respectively. Compared to V2, which employs the default point-to-plane ICP localization without degeneracy handling, V5 exhibits only a $1$--$2$ cm increase in lateral RMSE. This small performance gap indicates that the remaining error difference largely reflects a robustness-accuracy trade-off. Since the proposed system is designed to operate reliably in geometrically degenerate environments, we prioritize navigation robustness while incurring only a minimal loss of accuracy under feature-rich conditions. An adaptive degeneracy threshold could potentially further reduce the accuracy gap while maintaining robustness under varying levels of geometric degeneracy, which we leave for future work.}

\subsection{\revision{Runtime Performance}}\label{subsec:runtime}

\revision{
To evaluate the computational efficiency of the proposed pipeline and assess its viability for real-time operation, we analyze the execution times of all six pipeline variants introduced in Section~\ref{sec:exp_setup_env} (see Table~\ref{tab:tr_pipeline_variants}).
%
Runtimes are averaged on a per-frame basis over all field trials for each variant and reported in Table~\ref{tab:runtime_results}. For clarity, the individual module runtimes are grouped into four categories: \emph{Preprocessing}, comprising raw sensor data conversion (\texttt{aeva\_converter}), Doppler velocity estimation (\texttt{preprocessing\_doppler}), and either uniform (\texttt{preprocessing}) or curvature-based downsampling (\texttt{preprocessing\_curvature}); \emph{Odometry}, comprising either correspondence-free Doppler-inertial odometry (\texttt{odometry\_doppler}) or geometry-based ICP odometry (\texttt{odometry\_icp}); \emph{Localization}, comprising either our proposed CDA-ICP or standard point-to-plane ICP localization; and \emph{Other}, comprising overhead tasks required for pipeline execution, including map maintenance, submap recall, memory management, and vertex creation checks.

The results highlight the computational savings achieved by our proposed methodology. The total per-frame runtime of our proposed system (V5) is $40 \ \mathrm{ms}$, a $58\%$ reduction in computational cost compared to the default \ac{LTR} baseline (V1, $96 \ \mathrm{ms}$). The largest source of savings is the replacement of geometry-based ICP odometry and localization with Doppler-inertial odometry and CDA-ICP localization. In V1, standard ICP odometry and localization are computationally expensive due to iterative nearest-neighbour searches performed on uniformly downsampled point clouds, which retain a comparatively high point density. The proposed localization framework achieves substantial speedups over the baseline ($6 \ \mathrm{ms}$ in V5 vs.\ $33 \ \mathrm{ms}$ in V1). This is because curvature-based preprocessing selectively retains geometrically informative points, significantly reducing the point density used in scan-to-map matching. Comparing the localization module runtime in V3 ($9 \ \mathrm{ms}$), which uses curvature-based preprocessing with standard ICP localization, to V2 ($32 \ \mathrm{ms}$), which uses uniform preprocessing with standard ICP localization, isolates this effect: curvature-informed point selection alone reduces ICP localization runtime by approximately $72\%$ by retaining a smaller, more geometrically informative point set that accelerates convergence, independent of the CDA-ICP formulation itself.

Comparing the degeneracy-aware localization variants (V4--V6), which share the same curvature-based preprocessing and Doppler-inertial odometry, isolates the cost of the underlying degeneracy-handling formulation. V4, which uses solution remapping with fixed point-to-plane residuals, and V5, which extends this to adaptive point-to-point/point-to-plane residual weighting, achieve comparable localization runtimes (both $6 \ \mathrm{ms}$). V6, which replaces solution remapping with the constraint-based quadratic program of~\cite{tuna2025informed}, incurs a higher localization cost ($9 \ \mathrm{ms}$) due to the added overhead of solving a constrained optimization problem at each iteration. Despite this, all three variants (V4--V6) remain well within real-time requirements, with total per-frame runtimes of $40$, $40$, and $42 \ \mathrm{ms}$, respectively.

\begin{table}[!t]
    \centering
    \caption{\revision{Average runtime performance breakdown per frame (ms/call, rounded to the nearest millisecond) for the six pipeline variants. V1 is the default LT\&R baseline; V5 is the proposed pipeline.}}
    \label{tab:runtime_results}
    \begin{tabularx}{\linewidth}{X c c c c c c}
        \toprule
        \textbf{Module} & \textbf{V1} & \textbf{V2} & \textbf{V3} & \textbf{V4} & \textbf{V5} & \textbf{V6} \\
        \midrule
        Preprocessing            & 25             & 28             & 27             & 29             & 28             & 27             \\
        Odometry                 & 27             & 1              & 1              & 1              & 1              & 1              \\
        Localization             & 33             & 32             & 9              & 6              & 6              & 9              \\
        Other                    & 11             & 12             & 4              & 5              & 5              & 5              \\
        \midrule
        \textbf{Total Runtime (ms)}   & \textbf{96}    & \textbf{74}    & \textbf{41}    & \textbf{40}    & \textbf{40}    & \textbf{42}    \\
        \bottomrule
    \end{tabularx}
\end{table}
Overall, these results demonstrate that the proposed pipeline comfortably satisfies real-time requirements, and that the modest additional cost of curvature estimation and CDA-ICP localization is more than compensated by the reduction in downstream registration complexity relative to the default \ac{LTR} baseline.
}
\section{Conclusion}

This paper presents a degeneracy-resilient \ac{TR} navigation system using a \ac{FMCW} lidar for geometrically challenging environments. We extended a correspondence-free Doppler–inertial odometry method with pose uncertainty estimation, enabling principled uncertainty propagation into localization. To improve scan-to-map alignment under geometric degeneracy, \revision{we proposed a curvature-enhanced, degeneracy-aware ICP~(CDA-ICP) method that leverages per-point curvature for preprocessing, data association and adaptive registration formulation. The resulting optimization employs block scaling to perform degeneracy detection and optimization jointly in a unified scaled space, with the solution subsequently mapped back to the original space. Simulation and offline evaluations examined the proposed design choices, including curvature estimation, the registration formulation, degeneracy threshold sensitivity, and odometry uncertainty propagation.}

\revision{Extensive closed-loop field experiments were conducted to evaluate the complete system} across environments ranging from feature-rich to severely feature-limited. In structured environments, the system achieved tracking performance comparable to the baseline Doppler \ac{LTR} pipeline. In contrast, in a flat airport runway scenario, where ICP-based \ac{LTR} and other pipeline variants failed, the proposed system maintained reliable autonomous navigation, demonstrating robustness under extreme geometric degeneracy. By combining geometry-independent Doppler odometry with curvature-enhanced, degeneracy-aware localization, the system remains stable by updating well-constrained motion directions while relying on odometry in degenerate subspaces. \revision{Runtime evaluation further showed that the proposed pipeline maintains real-time operation with reduced computational cost compared with the default \ac{LTR} baseline. In summary, the proposed system enables reliable and computationally efficient autonomous navigation across environments with widely varying geometric structure, making it well suited for unstructured environments,} such as planetary surfaces, where strong geometric features cannot be assumed.


\bibliographystyle{./IEEEtranBST/IEEEtran}
\bibliography{./IEEEtranBST/IEEEabrv,./biblio}

\vfill

\end{document}